\PassOptionsToPackage{table}{xcolor}

\documentclass[acmsmall, anonymous=false]{acmart}

\usepackage{graphicx}
\usepackage{titlesec}
\usepackage{xcolor}
\usepackage{natbib}
\usepackage[ruled]{algorithm2e} 
\usepackage{etoolbox}
\usepackage[all]{nowidow}
\usepackage{subcaption}
\usepackage[utf8x]{inputenc}
\usepackage{textcomp}
\usepackage[export]{adjustbox}
\usepackage{listings}
\usepackage{wrapfig}
\usepackage{mwe}
\usepackage{caption}
\usepackage{subcaption}
\usepackage[width=.75\textwidth]{caption}
\usepackage{placeins}
\usepackage{array,tabularx,ragged2e,booktabs}
\usepackage{flafter} 
\usepackage{arydshln}
\newcolumntype{Y}{>{\centering\arraybackslash}X}
\newsavebox{\bigimage}
\usepackage{arydshln}

\AtBeginDocument{%
  \providecommand\BibTeX{{%
    \normalfont B\kern-0.5em{\scshape i\kern-0.25em b}\kern-0.8em\TeX}}}

\setcopyright{rightsretained}
\acmJournal{PACMHCI}
\acmYear{2021} \acmVolume{5} \acmNumber{CSCW2} \acmArticle{317} \acmMonth{10} \acmPrice{}\acmDOI{10.1145/3476058}

\acmYear{2021} 
\acmVolume{5} 
\acmNumber{CSCW2}
\acmArticle{317}
\acmMonth{10} 
\acmPrice{15.00}

\begin{document}

\title{Do Datasets Have Politics? Disciplinary Values in Computer Vision Dataset Development}


\author{Morgan Klaus Scheuerman}\authornote{This work was conducted while the first author was at Google Research.}
\email{morgan.scheuerman@colorado.edu}
\orcid{0000-0002-6049-3965}
\affiliation{%
  \institution{University of Colorado Boulder}
  \department{Department of Information Science}
  \state{CO}
  \country{USA}
}

\author{Emily Denton}
\email{dentone@google.com }
\orcid{}
\affiliation{
    \institution{Google Research}
    \country{USA}
}

\author{Alex Hanna}
\email{alexhanna@google.com}
\orcid{0000-0002-8957-0813}
\affiliation{
    \institution{Google Research}
    \country{USA}
}

\renewcommand{\shortauthors}{Morgan Klaus Scheuerman, Emily Denton, and Alex Hanna}

\begin{abstract}
  Data is a crucial component of machine learning. The field is reliant on data to train, validate, and test models. With increased technical capabilities, machine learning research has boomed in both academic and industry settings, and one major focus has been on computer vision. Computer vision is a popular domain of machine learning increasingly pertinent to real-world applications, from facial recognition in policing to object detection for autonomous vehicles. Given computer vision’s propensity to shape machine learning research and impact human life, we seek to understand disciplinary practices around dataset documentation—how data is collected, curated, annotated, and packaged into datasets for computer vision researchers and practitioners to use for model tuning and development. Specifically, we examine what dataset documentation communicates about the underlying values of vision data and the larger practices and goals of computer vision as a field. To conduct this study, we collected a corpus of about 500 computer vision datasets, from which we sampled 114 dataset publications across different vision tasks. Through both a structured and thematic content analysis, we document a number of values around accepted data practices, what makes desirable data, and the treatment of humans in the dataset construction process. We discuss how computer vision datasets authors value efficiency at the expense of care; universality at the expense of contextuality; impartiality at the expense of positionality; and model work at the expense of data work. Many of the silenced values we identify sit in opposition with social computing practices. We conclude with suggestions on how to better incorporate silenced values into the dataset creation and curation process.
\end{abstract}



\maketitle

\section{Introduction}
Data powers modern machine learning and artificial intelligence. Major tech corporations have built their intellectual and financial wealth by monetizing massive caches of text, images, transactions, and relationships. Modern nation-states and municipalities are increasingly data-driven, like when using administrative and court data to make decisions about provision of welfare benefits, who goes to jail, and whether to allow children to remain with their parents. In short, major institutions of life are awash in data and predictive technologies powered by machine learning.

Computer vision is a subfield of machine learning focused on analyzing and extracting information from images and videos to make predictions. The types of visual processing tasks that computer vision practitioners are concerned with are varied, ranging from the identification of animals and plants to the localization and tracking of people in surveillance footage. The stakes of computer vision research are incredibly high. Facial recognition, the task under the computer vision umbrella which has received the most attention, has been compared to plutonium, its harms far outweighing its benefits \cite{10.1145/3313129}; activists and researchers have already successfully lobbied for its banning in multiple jurisdictions \cite{selingerWhatHappensWhen2019}. But computer vision extends into other consequential domains, including autonomous vehicles, robotics, military target identification, refugee resettlement, and humanitarian aid. Given how modern computer vision works, to analyze the details of modern algorithms means that we must interrogate the datasets used to train, evaluate, and standardize these algorithms.

Modern computer vision research relies heavily on datasets of images and/or videos that are used to develop and evaluate computer vision algorithms. Broadly speaking, these datasets are composed of a collection of visual data with an associated set of annotations per data instance. A subset of the image annotations are conceptualized as target variables, variables a computer vision algorithm is designed to predict given an input. In this sense, computer vision datasets also instantiate computer vision tasks. For example, the facial analysis task of age prediction necessitates a dataset consisting of faces with associated age annotations. Computer vision uses a subset of data instances, known as training data, to fit the parameters of a model. Another subset of the data is known as testing data, leveraged to evaluate the model and estimate the ability of the model to generalize to images not seen during training. In addition to providing the technical backbone of computer vision algorithms, datasets are also leveraged by research communities with an explicit goal of advancing research and algorithmic developments on a particular problem; these are known as benchmark datasets. Benchmark datasets are often accompanied by standardized metrics to report model performance and to facilitate quantitative comparisons. In some instances, benchmark datasets are formalized in yearly challenges where researchers compete to develop the best performing model.

Given the centrality of datasets to computer vision practices, many researchers have proposed new data reporting practices to increase data transparency. For instance, \citeauthor{gebruDatasheetsDatasets2020}, \citeauthor{benderDataStatementsNatural2018}, and \citeauthor{hollandDatasetNutritionLabel2018} have all put forth data reporting frameworks for improving dataset documentation \cite{gebruDatasheetsDatasets2020, benderDataStatementsNatural2018,  hollandDatasetNutritionLabel2018}. This work is forward-looking, intended to improve future dataset practices. However, these guidelines do not generally offer guidance in incorporating specific \emph{values} into the dataset curation process, or even in articulating the values shaping dataset development. Existing retrospective critiques of the values of computer vision datasets have been constrained to specific identity categories, such as race \cite{buolamwiniGenderShadesIntersectional2018} and gender \cite{scheuermanHowComputersSeeGender2019} and their potential downstream impacts on model bias. Prior work has not examined the broader politics historically and presently incorporated into vision dataset development. Thus, we currently do not understand exactly what values are present, or absent, in computer vision data practices. Unlike in social computing and technology ethics work, computer vision\,---\,and machine learning more broadly\,---\,does not have a documented culture of reflexivity about values in their work \cite{rajiYouCanSit2021, joLessonsArchivesStrategies2020}. 

In failing to articulate the values that shape dataset development, dataset authors aid in rendering the value-laden components of the dataset invisible, whether intentionally or not. Moreover, when values inherent in dataset development\,---\,through task formulation, collection of data instances, structuring of data via annotation processes, and more\,---\,are left unaccounted for, dataset creators signal that the myriad of decisions were not important, or even consciously made. Consequently, the resulting dataset is more likely to be viewed as a natural reflection of the world, rather than a constructed and situated reflection of a particular worldview. The manner in which dataset developers choose to present and describe their work has tangible consequences for how their dataset is adopted and, more generally, impacts cultures of dataset development and use within the field. The lack of the aforementioned documentation practices are not solely on the shoulders of individual authors; it reflects larger institutional values within the field of computer vision, leading to an unchanging culture within conferences, journals, and education that encourages better documentation. A naturalized and objective practice contributes to a culture of uncritical and unquestioning dataset use by many computer vision practitioners. 

In this work we leverage the texts associated with computer vision datasets to examine the values operative in dataset development. How dataset creators choose to describe their datasets and the processes that went into their development signals what the creators value. These texts provide a unique grounds for analyzing the values underlying dataset development in the field. In this work, we focus on answering the following research questions:

\begin{enumerate}
    \item What do authors document about the dataset curation process?
    \item How do authors document the dataset curation process? What language do they employ in describing the process, the dataset itself, and its value to their audience?
    \item What do answers to questions 1 and 2 communicate about the values of computer vision datasets? What values are \emph{not} communicated?
\end{enumerate}

Specifically, we analyzed documentation from 113 different computer vision datasets (114 publications) across a variety of vision-related tasks\,---\,face-based, body-based, and non-corporeal tasks like object recognition. We built an extensive codebook to capture different segments of the dataset curation pipeline, from data collection to annotation to dissemination. We employed both structured content analysis and qualitative thematic analysis. Our structured analysis was focused on \emph{what} authors explicitly document in their dataset curation, in terms of what they feel is valuable to communicate to readers; our qualitative analysis was focused on excavating \emph{how} the process was communicated, in terms of what language and statements are used to communicate the dataset process and its contribution. Synthesizing both structured and thematic analysis allowed us to identify specific, overarching values in the computer vision dataset curation process. 

We present our findings in three themes, focused on the different levels of the dataset process. First, we present findings on the disciplinary-specific practices of dataset authors. Second, we present findings on data instances, and what makes certain data sought after for computer vision datasets. Third, we present findings on human actors involved in the data process\,---\,the annotators and data subjects\,---\,and how dataset authors discuss their roles. Through our discussion, we synthesize these three themes into larger values around the computer vision dataset curation process.

We identify and discuss four values of computer vision datasets: efficiency, universality, impartiality, and model work. For each present value, we identify a contrasting \emph{silenced} value\,---\,values that are overlooked or implicitly devalued in favor of the embraced values. Efficiency is valued over care, a slow and more thoughtful approach to dataset curation. Universality is valued over contextuality, a focus on more specific tasks, locations, or audiences. Impartiality is valued over positionality, an embracing of the social and political influences on understanding the world. And model work is valued over data work, with most authors focusing little on explicating data practices in favor of detailing the proposed machine learning method or model. For each silence, we recommend steps towards actively valuing them in dataset curation. In highlighting what values are currently embraced in computer vision data practices, and what values are systematically overlooked or devalued, we see opportunities for intervention throughout the dataset curation pipeline. We argue that embracing the silenced values have the potential to change the process of curating computer vision data to be more trustworthy, ethical, and human-centered. 

\section{Related Work}

\subsection{Designing for Values}
Scientific disciplines are encultured with their own specific practices and values. Philosophers of science have examined the relationship between practices and values of certain scientific disciplines for some time \cite{breeze2011disciplinary, becherDisciplinaryDiscourse1987, cooper1995representing}, as well as how those values are shaped by and shape social life \cite{winnerWhaleReactorSearch2020, goldmanSocialImpactAnalysis2020, salterEconomicBenefitsPublicly2001, smithMeasuringSocialImpact2001}. Foucault defined knowledge as a practice imbued with discursive power, in which a scientific ``truth'' is constructed through classificatory practices \cite{powell2015disciplining}. The intricate webs connecting science with social and political institutions shape how we see and interact with the world. Some sciences, which are particularly focused on exploring aspects of social life, have profound implications for the practices of individual and collective human life. Foucault’s analysis of sexuality \cite{foucault1990history}, Repo’s analysis of gender \cite{repo2015biopolitics}, and Stoler’s analysis of race \cite{stolerRaceEducationDesire1995} reveal historical genealogies in which science has intervened to shape how certain groups of people are viewed, treated, and interacted with\,---\,in everything from academic theory to family life to medicine. 

Scientific and technological production is necessarily an exercise of judgement, requiring a series of value judgements on what should be made visible and invisible, documented and undocumented. There is a broad literature within science and technology studies scholarship which focuses on the gap between what is documented and what gets executed in the laboratory. Classic lab studies highlight the operation of the gradual black boxing of technologies and techniques as they become more commonplace \cite{latour1987science, latour1986laboratory}. Others have highlighted the role of the formal organization and technoscientific practice \cite{vaughan1999role}, the social nature of accepting the experimental results of certain configurations of scientific instrumentation \cite{collins1992changing}, and scientific practice as being an extension of culture, historical contingency, and other material considerations \cite{pickering2010mangle}. We acknowledge and build on this work, focusing specifically on data practices.

Central to the practice of science---and as such, to machine learning---is data, the vague notion of ``raw'' facts, text, numbers, or other inputs that can be organized via analysis into meaningful information \cite{foxNotionDataIts1994, schoch:hal-00920254, wardUndefinedDataSurvey2013, gitelmanRawDataOxymoron2013a}. Like the practice of science broadly, data is imbued with specific values. The kinds of data collected, how it is collected, and how it is analyzed all reflect disciplinary and researcher values. \citeauthor{bowker2005memory} examined how the data considered most interesting and classifiable shapes how data is then stored, and how data deemed uninteresting or difficult to classify can become underspecified or entirely lost in larger data catalogues \cite{bowker2005memory}. Similarly, \citeauthor{vertesiValueDataConsidering2011} argue that the values in data arise in the nature of data production \cite{vertesiValueDataConsidering2011}. Viewing data as non-neutral is not a new concept.

Though defining data is difficult and contextual to the discipline in which it is situated, classification is a core tenant of analyzing, disseminating, and organizing data. How data, and information gleaned from data, is catalogued and applied is political. \citeauthor{sorting} discussed the politics imbued in technical classification systems, like those of the International Classification of Diseases (ICD) \cite{sorting}. Winner famously discussed the implications of the built environment and its political ramifications \cite{winnerWhaleReactorSearch2020}, while Suchman scrutinized how categorical decisions in system design could hold up specific social orders \cite{suchmanCategoriesHavePolitics1993}, and Agre criticized approaches to artificial intelligence for undervaluing critical social theories \cite{Agre2006TowardAC}. Value misalignments can often lead to artifact abandonment, or active resistance from the intended stakeholders (e.g. \cite{Davis1989, DeVito2017}). The acknowledgment of values in computing led Friedman to develop a value-sensitive approach to designing systems \cite{friedmanValuesensitiveDesign1996}, which has gone on to influence many design frameworks and approaches \cite{abebeRolesComputingSocial2020, sengersReflectiveDesign2005, blevisSustainableInteractionDesign2007, 10.1007/978-3-030-50334-5_1, https://doi.org/10.1002/pra2.2018.14505501013, Moreau2019}. \citeauthor{bardzellFeministHCITaking2010}, \citeauthor{costanza-chockDesignJusticeCommunityled2020}, \citeauthor{ogbonnaya-ogburuCriticalRaceTheory2020}, and \citeauthor{keyesHumanComputerInsurrectionNotes2019} have introduced other examples of social computing design frameworks which have explicitly embraced specific politics and values \cite{bardzellFeministHCITaking2010, costanza-chockDesignJusticeCommunityled2020, ogbonnaya-ogburuCriticalRaceTheory2020, keyesHumanComputerInsurrectionNotes2019}

Particularly with its focus on real-world applications, the artifacts produced by machine learning researchers and practitioners\,---\,for example, in the form of datasets\,---\,are value-laden \cite{hannaCriticalRaceMethodology2019, Scheuerman2020, slotaGoodSystems}. Given the necessity of data to modern computer vision research and applications, this work focuses specifically on examining the underlying values of computer vision datasets\,---\,the practices of collecting, classifying (often, through annotating), and disseminating data to be used by researchers and practitioners for building and evaluating computer vision models. While we do not employ a specific value-centered design framework to analyze computer vision datasets, we provide insight into the values already embedded into datasets and how they sit in opposition to other potential values.

\subsection{Critiques of CV Datasets and Dataset Development Practices}

The datasets underpinning modern computer vision algorithms have come under increased scrutiny in recent years, raising concerns regarding the image contents and categorical structures, as well as methods of construction, documentation, and maintenance. Recent audits of computer vision datasets have uncovered serious concerns regarding the degree and manner of representation of different socio-demographic groups. For example, \citeauthor{buolamwiniGenderShadesIntersectional2018} found darker-skinned women to be heavily underrepresented within a range of widely used facial analysis datasets \cite{buolamwiniGenderShadesIntersectional2018}. Object recognition datasets have a documented bias towards Western countries \cite{Vries_2019_CVPR_Workshops, shankarNoClassificationRepresentation2017a}. Stereotype aligned correlations between gender and the activities being depicted in images have also been identified in several computer vision datasets (e.g., overrepresenting women in images depicting cooking and shopping \cite{10.1007/978-3-030-01219-9_47, zhao-etal-2017-men}). A recent audit of ImageNet found the dataset contained significant gender biases, and even the inclusion of non-consensual pornographic imagery, depicting predominantly women \cite{prabhuLargeImageDatasets2020}. 

The categories structuring computer vision datasets have also been critiqued recently. In an examination of the `person' categories within ImageNet\,---\,derived from the WordNet hierarchy \cite{fellbaumWordNet2012}\,---\,\citeauthor{crawford2019excavating} found the inclusion of misogynistic terms, racial slurs, and otherwise offensive labels \cite{crawford2019excavating}.\footnote{Following the release of Crawford and Paglen's article, the ImageNet creators removed a subset of the person categories from the dataset.} \citeauthor{prabhuLargeImageDatasets2020} extended this analysis to other image datasets that have derived their categorical structure from WordNet, and found the TinyImages dataset also contained slurs and other offensive labels \cite{prabhuLargeImageDatasets2020}.\footnote{TinyImages has since been removed from the web.} \citeauthor{Scheuerman2020} raised concerns regarding the use of gender and racial categories in facial analysis datasets, finding that dataset creators tend to present these classifications as natural or immutable characteristics \cite{Scheuerman2020}. At the level of data classifications more broadly, \citeauthor{senTurkersScholarsArafat2015} problematize the broad use of so-called ``universal gold standards'' in benchmark datasets at all \cite{senTurkersScholarsArafat2015}.

Examinations of the practices of dataset development have exposed the widespread devaluation of data work. \citeauthor{sambasivanEveryoneWantsModel2021} found high-quality dataset development to be one of the most undervalued components of machine learning practice \cite{sambasivanEveryoneWantsModel2021}. Dataset development is rarely glamorized like algorithmic developments, which is reflected in peer review processes that make it difficult to publish work focusing exclusively on datasets \cite{heinzerling2019cleverhans}. Furthermore, dataset development is often omitted entirely from machine learning curriculums and textbooks (e.g., \cite{Goodfellow2016}). \citeauthor{joLessonsArchivesStrategies2020} characterize the resulting culture of dataset development as one that embodies a \textit{laissez-faire} attitude, which they contrast with the careful and critical curatorial practices of archivists \cite{joLessonsArchivesStrategies2020}.  \citeauthor{paulladaDataItsDis2020} discuss how dataset culture within machine learning prioritizes speed for the achievement of algorithmic performance on a fixed set of benchmarks with little regard to the implications of data reuse, data management, and legal issues \cite{paulladaDataItsDis2020}. Further, privacy, consent, and licensing concerns permeate various stages of computer vision dataset development \cite{dulhantychrisIssuesComputerVision2020, khanHanna2020}. Many of the most commonly used large-scale image datasets have been developed by scraping images from the web, through search engines and photo-sharing websites such as Flickr, which has raised the alarm from privacy advocates \cite{murgiamadhumitaWhoUsingYour2019}.  

Our work builds on this growing body of research examining computer vision datasets and their associated practices, but can be distinguished from previous work in several ways. First, we present the first large-scale systematic study of the documentation associated with computer vision datasets. Previous critical studies of computer vision datasets have been localized in scope to collections of datasets associated with certain tasks or categories (e.g., \cite{prabhuLargeImageDatasets2020, crawford2019excavating}) or in-depth audits of a handful of commercial models or datasets (e.g., \cite{buolamwiniGenderShadesIntersectional2018, prabhuLargeImageDatasets2020}). Second, our work is the first to explicitly examine the shared disciplinary values underlying dataset development in computer vision. Understanding current values in vision datasets will aid the development of new dataset approaches that better account for values. 

\subsection{Dataset Documentation in Machine Learning}

Stemming in part from the lack of transparency of both the provenance and the contents of many machine learning datasets, several dataset documentation frameworks have been proposed in recent years. These different proposals have varied goals and stem from different academic communities, with many different monikers: datasheets, data statements, dataset nutrition labels, and dataset requirement documents. However, they are united in understanding the different ways that dataset development can affect the outcomes of machine learning systems. 

Gebru et al. take the inspiration for their framework, datasheets for datasets, from datasheets in the electronics industry. The authors provide a long list of questions to ask of each dataset, including motivations, composition of the dataset, the data collection process, the preprocessing and labeling processes, and the use and distribution of the data \cite{gebruDatasheetsDatasets2020}. \citeauthor{hollandDatasetNutritionLabel2018} riff off the nutritional label used to report facts about the nutritional value of foods, and provide a web tool to facilitate the creation of data nutritional labels \cite{hollandDatasetNutritionLabel2018}. \citeauthor{benderDataStatementsNatural2018} propose a similar documentation method: data statements, which is documentation specifically geared towards natural language processing (NLP) datasets \cite{benderDataStatementsNatural2018}. Drawing on value-sensitive design \cite{friedmanValuesensitiveDesign1996}, they compel NLP data authors to include language variety, speaker and annotator characteristics (such as ``presence of disordered speech,'' ``native language,'' and ``training in linguistics,'') speech situation, text characteristics, and recording quality. \citeauthor{Geiger2020}, meanwhile, do not provide a formal diagnostic or checklist for the construction of dataset documentation but form one implicitly by analyzing a set of papers focused on social computing \cite{Geiger2020}. Coming from the tradition of structured content analysis from the social sciences (e.g. \cite{krippendorff2018content}), the authors code papers for several items, including whether the data used human annotation, if they had come from in-house or crowdsourced annotators, if the annotators were compensated, if they had training and if the instructions are available, if they used an interrater reliability metric, and if the dataset is available. They found that most of this information is not available for the datasets reported. \citeauthor{hutchinsonAccountabilityMachineLearning2020} have proposed a data reporting framework which follows engineering principles of iteratively creating design, requirement, and maintenance documentation with the participation of many stakeholders across the data lifecycle \cite{hutchinsonAccountabilityMachineLearning2020}. \citeauthor{miceliDocumentingComputerVision2021} and \citeauthor{afzalDataReadinessReport2020} provide summarizations of each framework's defining characteristics \cite{miceliDocumentingComputerVision2021, afzalDataReadinessReport2020}. 

Our work is similar in spirit to the aforementioned dataset documentation proposals. However, with the exception of \citeauthor{Geiger2020}, nearly all of these publications propose frameworks which are aimed at adoption by stakeholders intending to build new datasets. In contrast, our work focuses retrospectively on examining current dataset documentation practices and the values structuring dataset development within computer vision. While prospective analytics aimed at influencing future data practice are urgently needed, many of the existing frameworks don't fully account for current data practices. As noted by \citeauthor{hutchinsonAccountabilityMachineLearning2020} and \citeauthor{sambasivanEveryoneWantsModel2021} above, as well as \citeauthor{NIPS2015_86df7dcf} \cite{NIPS2015_86df7dcf}, data work is largely erased, despite its importance to machine learning practices. Although myriad tools and techniques exist for exploring data after being received from a single source (e.g., \cite{mullerInterrogatingDataScience2020}), the practices that orient how and which data are collected are less understood. Thus, studying data practices from existing dataset artifacts becomes critically informative. Further, with the exception of \cite{benderDataStatementsNatural2018}, which is oriented specifically around NLP dataset development, existing dataset documentation frameworks do not provide clear guidance for explicitly articulating values. Transparency without a normative orientation, or explicitly outlining one's values, implies the dataset reflects a ``view from nowhere,'' which runs counter to thought in human-computer interaction and feminist studies of technoscience \cite{haraway1988situated}. Our systematic analysis of the documentation associated with computer vision datasets provides much needed insight into what computer vision practitioners value in dataset practice, and what dataset curators are attentive to during dataset development. 

\section{Methods}
\subsection{Researcher Positionality}

Positionality statements situate the context from which the research was conducted and the authors have conducted and interpreted it \cite{attiaBeComIng2017}. All three authors work within the computing space, particularly on issues of fairness and equity in machine learning. The first author has a background in human-computer interaction, as well as media studies and gender studies; the second author has a background in computer science, specifically machine learning and computer vision; and the third author has a background in sociology. All three authors conducted their work while at Google.

We realize that our own perspectives color how we are interpreting these values, and that computer vision researchers may not characterize their values in the same ways. This may be seen as a weakness of the study \,---\, that because researchers from the computer sciences do not interpret their values as such, our results may lack external validity. However, we don't believe that this is a weakness of our study, but a strength. Because we explicitly acknowledge our own positionality in this analysis, we hope to remain reflexive in acknowledging our own subject position as researchers concerned with social computing and the workings of AI as sociotechnical systems. Moreover, we highlight a perspective that is oriented more towards science as both a technical {\it and} social practice.

\subsection{Manuscript Corpus and Keyword List}
We identified relevant computer vision datasets by examining the datasets used for a wide variety of computer vision tasks in academic papers. To obtain a corpus of manuscripts, we used IEEE proceedings, which is the publisher of most of the premier computer vision conferences (e.g., Computer Vision and Pattern Recognition (CVPR)). We identified computer vision proceedings by (1) selecting proceedings related to computer vision using the IEEE proceedings list; and (2) broadly searching “computer vision” in IEEE Xplore, the organization's digital library. By employing both these search methods, we were able to triangulate on potentially missed manuscripts. This method yielded a corpus of 50,694 computer vision manuscripts with metadata. 

Before examining the manuscripts for datasets, we first needed to define a set of computer vision tasks that we could narrow our search to. By task, we mean the broad classification of a problem that a computer vision is meant to solve (e.g., facial recognition, object detection, pedestrian detection, etc.), which may be associated with a specific type of data (e.g., face images). Given the vast diversity of computer vision literature, selecting datasets by task ensured a diverse variety of datasets meant for different purposes. To define a list of tasks, we parsed the keyword metadata in our corpus. We used both standardized keywords (IEEE keywords) and personalized keywords (author keywords) to reduce the risk of gaps in our task list. This gave us 247,126 keywords with which to work. To make the keyword list more manageable for analysis, we narrowed to keywords used 50 or more times in the corpus. This gave us a list of 1,622 keywords, representing about 57\% of all keywords in the list. We then manually removed keywords that were too high-level and not task-specific (e.g., computer vision) or were too vague to be tied to any tasks (e.g., video, engines). This left us with a manageable list of 345 keywords. During our analysis, we also organically derived 10 keywords from abstracts while checking for understanding of some vague keywords (e.g., gunshot detection). Our finalized keyword list was 355 keywords.

These finalized keywords represented both tasks and data types. We thematically clustered the keywords from our list into 21 conceptually related topics (e.g., medical tasks, low-level image processing tasks, image generation tasks, etc.). We then reviewed these 21 concepts as a team to determine which clusters were broadly scoped to “any type of data” (e.g., image processing, image indexing, image enhancement, etc.). We removed 12 clusters of keywords during this phase. This resulted in nine conceptual clusters of 130 keywords that we then grouped into three broad categories: Face-Based, Body-Based, and Non-Corporeal (as in, not related to human bodies). We grouped keywords into related categories of tasks under each of the three clusters (e.g., the category ``gender classification'' contained the keywords ``gender classification,'' ``gender estimation,'' ``gender prediction,'' and ``gender recognition''). 

\subsection{Identifying Computer Vision Datasets}

We went back to the original manuscript corpus and randomly sampled five papers for each task category (e.g., for gender classification). We used the keywords in each category to sample by keyword in the corpus. For example, we randomly sampled five papers that matched the keyword ``body detection,'' and so on. If we could not identify five papers using a single keyword, or the papers did not contain dataset references, we cycled through keywords in each task category and resampled. For each random sample, we read the paper to determine which dataset(s) were being used or cited. We listed all datasets found within each paper under that task keyword. We gathered 331 unique datasets from this process.

After sampling each keyword in our task categories, we then found each original paper associated with the dataset. While doing this, we also snowballed new datasets we found in two ways: (1) we found a new dataset reference in the original dataset publication; or (2) we found a new dataset from the associated authors or organization. We snowballed 355 additional datasets. We did not snowball new datasets from online lists, forums, or other sources where we could not verify the methodology. We did, however, supplement face datasets from the open source list published by \citeauthor{Scheuerman2020}, which provided an additional 66 datasets\footnote{\url{https://zenodo.org/record/3735400\#.YAHWk5NKjUJ}}. This provided us with a total of 487 datsets.\footnote{ In the process of conducting our coding, we also identified an additional 271 datasets, which we added to our dataset list, but were not part of the original population from which our sample was drawn, giving us a total of 753 datasets. All datasets in the corpus are provided at \url{https://zenodo.org/record/4613146\#.YJwdwKhKiF5}.} 

\subsection{Sampling Datasets for Analysis}

For conducting our analysis, we decided to sample a more manageable number of datasets from our full corpus of 487. We sampled in two ways. First, given we felt it was important to understand the most popular datasets, we sampled datasets with over 4,000 citations on Google Scholar. This yielded 13 datasets that were a mix of Face-Based, Body-Based, and Non-Corporeal. Given the size and vast documentation of ImageNet, as well as its influence on the field and large number of citations per paper, we decided to code each major ImageNet paper separately. Second, we sampled 100 additional datasets stratified by each category (face, body, and non-corporeal). We decided to sample each category proportionally to the number in the overall corpus, rather than equally across all categories. We chose this approach to reflect the implicit popularity of certain types of datasets in the computer vision community. Our final sample included 113 datasets: 47 face-based, 25 body-based, and 41 non-corporeal. 

\begin{table}[hbt!]
\centering
\def\tabularxcolumn#1{m{#1}}
\begin{tabularx}{\linewidth}{YYYY}
        \hline
        \rowcolor[HTML]{EFEFEF} 
        \multicolumn{4}{c}{\cellcolor[HTML]{EFEFEF}\textbf{Corpus Breakdown}} 
        \\ 
        \hline
        \textit{\textbf{Property}} 
        & \multicolumn{1}{c}         
        {\textit{\textbf{Population}}}
        & \multicolumn{1}{c}
        {\textit{\textbf{Sample}}} 
        \\
        \cmidrule(lr){1-1}
        \cmidrule(lr){2-2}
        \cmidrule(lr){3-3}
        \cmidrule(lr){4-4}
        \rowcolor[HTML]{EFEFEF} 
         \textit{Face-Based} & 205 & 47
         \\
         \textit{Non-Corporeal}& 174 & 41
         \\ 
        \rowcolor[HTML]{EFEFEF} 
         \textit{Body-Based}& 105 & 25
         \\
          \textit{Citation mean / median \textsuperscript{a,b}}& 421.17 / 165 & 390.75 / 166
         \\  \rowcolor[HTML]{EFEFEF} 
         \textit{Year mean / median \textsuperscript{b} (rounded)}& 2011 / 2012 & 2011 / 2011
         \\
         \hline 
        \multicolumn{4}{c}{\textsuperscript{a}\footnotesize{Minus top 14 papers. }  \textsuperscript{b} \footnotesize{Only databases which have Google Scholar citation information.}} 
         \\
        \centering
     \end{tabularx}
  \caption[A table showing the breakdown of our corpus into categories and sampling statistics.]{\small A table showing the breakdown of our corpus into categories and sampling statistics.}
    \label{table:corpus}    
    \vspace{-5mm}
    
\end{table}
\FloatBarrier

Table \ref{table:corpus} displays descriptive statistics for the sample and the population corpus. Each broad category is proportionally represented in the sample. The mean year (rounded) of the population and sample are both 2011, and the median as 2012 and 2011, respectively. The range of years in our sample ranged from 1994 to 2020.
The sample appears to be a good representation of the types of papers which are common in the population corpus.

\subsection{Codebook Development}

Our focus was on the disciplinary practices of documenting the creation and maintenance of computer vision datasets. Therefore, we developed a codebook for comprehensively capturing different stages of the dataset curation process, from motivations to annotations to the availability of the data. The documentation we coded included research publications, the datasets themselves, websites, and auxiliary materials like slide decks. We developed the codebook iteratively during the structured analysis phase of our coding, often discovering through reading documentation or discussion among the research team new variables to capture. We met weekly to discuss disagreements and edge cases, which resulted in new category creation (e.g. new questions arose on whether the dataset used synthetic humans rather than real ones, as evidenced from our encounter with data created from generative machine learning algorithms). In the end, we coded for 95 different variables. The codebook is part of a larger research project aimed at understanding the genealogy of datasets \cite{Denton2020BringingTP}. For the purposes of this work, we focus on presenting findings which explicitly or implicitly spoke to the values imbued in datasets. Details on access to our full codebook and coded dataset can be found in Section \ref{access}. 

\subsection{Analysis}

\subsubsection{Structured Coding} 

We took both structured and thematic coding approaches for the analysis of our sample. Our structured content analysis focused on identifying common themes from the literature around machine learning datasets, including descriptions of data annotators \cite{irani2013turkopticon, gray2019ghost}, data availability for research purposes \cite{borgman2016big, pasquettoReuseScientificData2017}, properties of data subjects and categories \cite{Scheuerman2020}, and descriptions of decisions more generally which typically are obviated or obscured in the data collection process \cite{gebruDatasheetsDatasets2020, Geiger2020, benderDataStatementsNatural2018}. We found that using both structured and thematic coding of the datasets allowed us to abductively reason about the different dimensions, motivations, and silences around dataset creation that either method alone would not robustly allow.

Instead of approaching our structured content analysis through a process of independently coding data \cite{krippendorff2018content}, the authors split up the responsibility of coding each of the datasets, with the majority of the coding being performed by the first author. We decided against using a formal inter-rater reliability metric because coding articles for variables involved understanding the process of developing computer vision documents, and thereby could be disputed when looking at individual instances, which is one of the cases highlighted as a reason not to seek out an inter-rater reliability metric \cite{mcdonaldReliabilityInterraterReliability2019}. Instead, after coding data independently, we performed cross-check coding on each authors’ codings. Every author coded a random five datasets from one another and then met to discuss and resolve disagreements. We also went through the dataset entirely at the end of the project and validated the structured content by calculating cross tabulations in Python for variables which depended on each other for consistency. For instance, structured variables which relied on containing human subjects should have only been coded as Yes or No if the "Contains Human Subjects" variable had been set to Yes for the variable. 

\subsubsection{Thematic Coding} 

In addition to the structured content analysis, we also thematically coded language that communicated underlying values. Dataset documentation signals what the authors found necessary to communicate to the readers and, more broadly and in aggregate, what the computer vision community finds valuable to communicate about data. Similarly, what was not communicated about the dataset in documentation, even if present in the data itself, signaled what is valued and not valued. The structured coding phase of our analysis informed how we defined values, as every author became deeply familiar with the data and patterns in how datasets authors document their processes. Specifically, we define value language as statements the dataset authors imbued, often implicitly, with perspectives on importance and moral judgements. For instance, in the example of one the papers developed around ImageNet, the authors use the word \emph{accurate} to denote desirable properties of the dataset: 

\begin{quote}
    \emph{ILSVRC makes extensive use of Amazon Mechanical Turk to obtain \textbf{accurate annotations} … To collect \textbf{a highly accurate dataset}, we rely on humans to verify each candidate image collected in the previous step for a given synset.} \,---\,ImageNet (Paper 2) 
\end{quote}

We split out all value statements in each dataset and then performed an initial line-by-line open coding phase. After completing the open coding phase, we then performed a second more focused coding on each of the statements, grouping them into higher-level themes (e.g., the variety of open codes that could be grouped under the theme of “unbiased data”). We also developed categorical relationships between these higher-level themes (e.g., unbiased data and realistic data were what we defined as “desirable data properties”). We concluded our thematic coding phase by writing memos on each of the higher-level themes and their categorical relationships, and discussing and refining those memos as a team \cite{charmaz2006constructing}. 

\subsection{Access to Research Materials}
\label{access}

Given that computer vision datasets and their associated publications are largely available to the public, we felt it ethically responsible and appropriate to release our corpus of datasets. We have created an open access repository for: (1) our codebook of 114 coded datasets; (2) our population corpus of computer vision datasets; (3) our documentation on our coding procedures; and (4) our Python analysis code. For each of the datasets, we listed the original publication, venue it was published at, year published, and Google Scholar citation count at the time of collection. We encourage future researchers to use these resources for additional research or to build on our work. The data can be found at \url{https://zenodo.org/record/4613146#.YJwdwKhKiF5}.

\section{Findings}
We summarize the fourteen focused codes constructed from our qualitative content analysis and the broad themes under which they fall: \emph{dataset authors and their disciplinary practices, data and its data properties,} and \emph{human actors as annotators and data subjects}. Our codes are not mutually exclusive; some of them are highly related.

\subsection{Dataset Authors and their Disciplinary Practices}
\subsubsection{Data as Essential for Scientific Progress}
\label{data-as-essential}

Several of the datasets we sampled (18 of 114; 15.8\%) describe data as being crucial to particular subfields progressing beyond their current state-of-the-art. Although most of the papers we evaluated discussed some kind of new method or algorithm in concert with the release of the dataset (15 papers were dedicated entirely to the documentation of the dataset), the data typically worked in service of making progress to the field. This is because data can present new challenges, define a new task, or build a common research agenda for improving performance on a task. For instance, the authors of the Lippmann2000 dataset write that creating benchmark datasets allows standardization and mitigates the need for practitioners to create their own datasets:

\begin{quote}
    \emph{``These images have \textbf{accelerated advancement in the field} by, first, allowing scientists and engineers practicing at or near the state-of-the-art to carry out their work \textbf{without the additional burden of needing to become experts in generating quality images}; and, second, creating a small level of \textbf{ad hoc standardization} such that processed images are more quickly evaluated due to general familiarity with the original input.''} \,---\,Lippmann2000
\end{quote}

Similarly, many present the lack of relevant data to be one of the limiting factors in driving research in their subfield, and thus present their dataset as contributing to improving research in that subfield. The act of creating a dataset was both a barrier to entry for new work on a given task, but also an attempt to introduce a standard around the underresearched task. Data is difficult to attain, and it's attainment is a necessity of the machine learning process, including standards-making. The authors of Leeds Sports Pose introduce data to improve pose estimation:

\begin{quote}
    \emph{``As noted, current methods have been \textbf{limited by the lack of available training data} – to overcome this we introduce a new annotated dataset of 2,000 diverse and challenging consumer images which will be made publicly available.''} \,---\,Leeds Sports Pose
\end{quote}

\begin{figure*}[htb!] 
    \centering
    \textbf{Distribution of Proportions of Papers Dedicated to Dataset Documentation}\par\medskip
    \includegraphics[width=\textwidth]{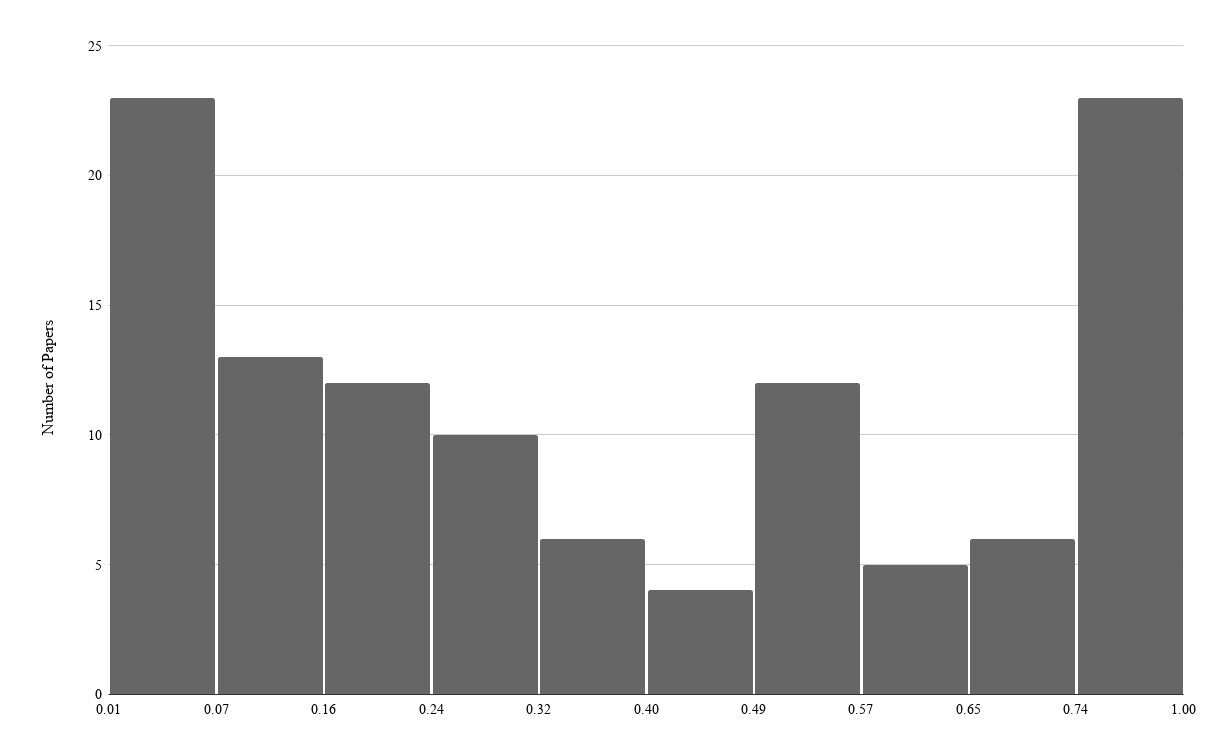}
    \caption[A histogram showing the distribution of the proportion that papers in the corpus dedicated to documenting only the dataset.]{\small A histogram showing the distribution of the proportion that papers in the corpus dedicated to documenting only the dataset. Proportion was calculated by hand-counting paragraphs in each paper. The histogram is bimodal, with the majority of papers having either (a) near-0\% of the paper about the dataset or (b) near-100\% of the paper about the dataset.}
    \label{proportion}
    \Description[A gray-scale histogram showing the distribution of the proportion that papers in the corpus dedicated to documenting only the dataset. Proportion was calculated by hand-counting paragraphs in each paper. The histogram is bimodal, with the majority of papers having either (a) near-0\% of the paper about the dataset or (b) near-100\% of the paper about the dataset.]{A gray-scale histogram showing the distribution of the proportion that papers in the corpus dedicated to documenting only the dataset. Proportion was calculated by hand-counting paragraphs in each paper. The histogram is bimodal, with the majority of papers having either (a) near-0\% of the paper about the dataset or (b) near-100\% of the paper about the dataset.}
\end{figure*}

However, some of these overt claims around the necessity of data to progress are not reflected in the actual practices of data management and curation used by researchers. The dataset itself was the main explicit contribution for a slim majority of papers (59 of 114 papers; 51.8\%), while a method or algorithm was the main contribution of the dataset in 53 of 114 papers (46.5\%) or an empirical study in 2 of 114 papers (1.8\%). Still, the dataset is rarely unaccompanied by a methodological innovation: papers typically contain some kind of new algorithm for the computer vision task under question (97 of 114 papers; 85.1\%). Lastly, we looked at how much of the paper was dedicated to describing the dataset. We calculated this value by counting the number of paragraphs dedicated to describing the dataset over the total number of paragraphs in the paper. Figure \ref{proportion} shows a bimodal distribution of how much of the paper is dedicated to describing the corpus, with a mean at 0.41 and a standard deviation of 0.33. This seems to indicate that papers are either wholly dedicated to the description of the dataset, or they provide scant information on the dataset, opting to discuss methodological innovations instead. Given the majority of papers were archival publications (105 of 114; 92\%), the lack of documentation in a paper about a dataset occurred even in archival publications --- publications in professional venues like conferences and journals.

\FloatBarrier

\subsubsection{Standardization for Evaluation and Reproducibility}
Some dataset authors (BigHand2.2M; KAIST; UMass FDDB; 300-W; IUPR) noted that they needed a common set of evaluation benchmarks, because the evaluation criteria which had been used in their subfield had too many different types of quantitative evaluations and needed a standardized benchmark. For example, the authors of 300-W were motivated to create their dataset to establish a standardized benchmark:

\begin{quote}
    \emph{``The main goal of this challenge is to compare the performance of different methods on a new-collected dataset using the same evaluation protocol and the same mark-up and hence to develop the first \textbf{standardized benchmark} for facial landmark localization.''} \,---\,300 Faces in the Wild
\end{quote}

``Standardized'' takes on a specific meaning for dataset authors. Generally, that meaning is one of quantitative measures which are viewed as reliable or objective. The authors of the IUPR dataset describe the impetus of benchmark datasets as being more ``objective'' as a ground for machine learning datasets: 

\begin{quote}
    \emph{``Ground-truth datasets are crucial for \textbf{objectively measuring the performance of algorithms} in many fields of computer science. The availability of such datasets for use in research and development lays the basis for comparative evaluation of algorithms.''} \,---\,IUPR
\end{quote}

Given the importance of data to standardizing algorithmic performance, dataset creators often claim to release datasets for the purposes of reproducibility and replicability, noting that this has been a failing of methods reporting results in the past. The authors of UG\^{}2 highlight that research would benefit from reproducible standards in datasets:

\begin{quote}
    \emph{`` new video benchmark dataset representing both ideal conditions and common aerial image artifacts, which we make available to facilitate new research and to simplify the \textbf{reproducibility of experimentation}.''} \,---\,UG\^{}2 Challenge Dataset
\end{quote}

Dataset authors also devalued qualitative or heuristic assessments of data quality and classifier results. Qualitative assessments were seen as less standardized and therefore less reliable. For example, the authors of HumanEva write that qualitative assessments decrease certainty and make it difficult to rigorously compare methods:

\begin{quote}
    \emph{``Despite clear advances in the field, evaluation of these methods remains mostly \textbf{heuristic and qualitative}. As a result, it is difficult to evaluate the current state of the art \textbf{with any certainty or even to compare different methods with any rigor}.''} \,---\,HumanEva
\end{quote}

Overall, dataset authors valued standardization, in terms of quantitative measurements, because they viewed those standardizations as objective, reliable, and reproducible.

\subsubsection{Open Source Data}
The majority the datasets which we sampled provided a URL or some web identifier for obtaining the dataset. 69 of 114 (60.5\%) of the datasets provide a URL in the paper; in addition, we were able to discover 28 additional websites via a web search of the publication or the dataset name, bringing the total number of sites were able find to 97 of 114 (85\%). Several authors stated this explicitly in their documentation (Abstract Paintings / Artistic Photographs; KinFaceW; IterNet RGB-D; NWPU-RESISC45; SFEW; Urban Stereo Scene; UvA-NEMO Smile; UG\^{}2). For instance, the authors of the UvA-NEMO Smile Database state that:

\begin{quote}
    \emph{``The database, its evaluation protocols and annotations are \textbf{made available to the research community}.''} \,---\,UvA-NEMO Smile Database
\end{quote}

Other authors (ASLLRP SignStream; CUAVE; SFEW) directly mention the research medium through which their data will be distributed, highlighting its benefit to the research community:

\begin{quote}
    \emph{``Finally, one of the main purposes of all speech corpora is to allow the comparison of methods and results in order to stimulate research and fuel advances in speech processing. This is a main consideration of the CUAVE database, \textbf{easily distributable on one DVD}.''} \,---\,CUAVE
\end{quote}

While most of the datasets report an URL (or have a findable URL through web search), many of the datasets did not have any institutional mechanism for the stability of the datasets. Despite the expressed value of open source datasets to the research community, most of the datasets are not maintained in a stable repository. Only 3 of the 114 (2.6\%) of the datasets attached a Digital Object Identifier (DOI) to them, while only one (0.9\%) was posted on an institutional repository such as Dataverse or Zenodo\,---\,The Child Affective Facial Expression (CAFE), hosted on NYU's Databrary (https://databrary.org/). Stability of these identifiers and hosting of datasets matter: of the 69 datasets which had a URL in the paper, only 46 (66.7\%) are still available. Of the 80 datasets in which data was openly accessible (without signing a user agreement, agreeing to a Terms of Service, or downloading a software package), 59 (73.8\%) were still downloadable. Table \ref{table:data-availability} summarizes full details about data hosting and availability.

\begin{table}[hbt!]
\centering
\def\tabularxcolumn#1{m{#1}}
\begin{tabularx}{\linewidth}{YYYY}
        \hline
        \rowcolor[HTML]{EFEFEF} 
        \multicolumn{4}{c}{\cellcolor[HTML]{EFEFEF}\textbf{Dataset Availability via Publication Documentation}} 
        \\ 
        \hline
        \textit{\textbf{Dataset Property}} 
        & \multicolumn{1}{c}         
        {\textit{\textbf{k}}}
        & \multicolumn{1}{c}
        {\textit{\textbf{N}}} 
        & \multicolumn{1}{c}
        {\textit{\textbf{\%}}} 
        \\
        \cmidrule(lr){1-1}
        \cmidrule(lr){2-2}
        \cmidrule(lr){3-3}
        \cmidrule(lr){4-4}
        \rowcolor[HTML]{EFEFEF} 
         \textit{URL in Paper} & 69 & 114 & 60.5
         \\
         \textit{Any Website (In Paper + Discovered through Search)} & 97 & 114 & 85
         \\ 
        \rowcolor[HTML]{EFEFEF} 
         \textit{Website in Paper Still Available} & 46 & 69 & 66.7
         \\
          \textit{Data Still Downloadable} & 59 & 80\textsuperscript{a} & 73.8
         \\ 
          \rowcolor[HTML]{EFEFEF} 
         \textit{DOI} & 3 & 114 & 2.6
         \\
         \textit{Hosted on Personal/Lab Website} & 102 & 114 & 89.5 
         \\ 
         \rowcolor[HTML]{EFEFEF} 
         \textit{Hosted on Institutional Repository} & 1 & 114 & 0.9
         \\
         \hline 
        \multicolumn{4}{c}{\textsuperscript{a}\footnotesize{The number of datasets which did not require registration to download.}}         
        \centering
        \end{tabularx}
  \caption[A table showing the breakdown of our corpus into categories and sampling statistics. NOTE: ImageNet is still treated as two separate publications in this table.]{\small A table showing the breakdown of our corpus into categories and sampling statistics.}
    \label{table:data-availability}    
    \vspace{-5mm}
\end{table}
\FloatBarrier

\subsubsection{Technical Documentation}
In many cases, the amount of documentation for the technical setup of data collection and algorithm design far outstripped the amount of documentation available for the actual data collection work and annotation and labeling procedures (see Figure \ref{proportion}). The authors of the KAIST Multi-Spectral Day/Night Data Set, created for autonomous vehicle research, provide a good example of this kind of technical detail in the configuration of lens capture devices placed on top of data collection cars:

\begin{quote}
    \emph{``In our case, we select a long focal lens to observe remote objects. If wanting to a wider field of view, a short focal lens may suffice. Recently, Tesla and Mobileye devised trifocal camera system (HoV−20°, −50°, −150°) as a new hardware configuration...''} \,---\,KAIST Multi-Spectral Day/Night Data Set for Autonomous and Assisted Driving
\end{quote}

The authors of the CMU-MultiPIE dataset, used for facial recognition research, report on data collection hardware, illumination configuration, and requisite computing hardware:

\begin{quote}
    \emph{``To systematically capture images with varying poses and illuminations during data acquisition we used a system of 15 cameras and 18 flashes connected to a set of Linux PCs. An additional computer was used as master to communicate with the independent recording clients running in parallel on the data capture PCs.''} \,---\,CMU-MultiPIE
\end{quote}

Reporting the technical details of hardware is ostensibly done to report improvements on data collection procedures, to resolve disagreements about instrumentation for other researchers working in this space, or to report on more precise conditions needed for future reproducibility of the experiment and data generation. 

\subsection{Data and its Desirable Properties}

\subsubsection{Diverse and Varied Data}
\label{diverse-data}
Common dimensions of diversity referenced by dataset creators include the diversity of scenes and object categories (ImageNet (Paper 1); KAIST; UG\^{}2; Scene Geometry Layout,;GHIM-10k); diversity of conditions of image capture, such as sensor quality, camera angle, or lighting conditions (TUM GAID; IUPR; Leeds Sports Pose; Exact Street2Shop); diversity of object poses or background clutter (IUPR; NWPU-RESISC45; ImageNet (Paper 1); Animals on the Web; INRIA Pedestrian; BSDS300); and diversity of the expressions or poses of individuals in the images (Exact Street2Shop). Diversity of camera quality was especially important for datasets that were motivated by real-world applications, where low-quality sensors would likely be common\,---\,most commonly, in intended surveillance applications (SCFACE; ND-QO-Flip; 300-W). 
Dataset creators emphasized the importance of diverse training data for ensuring model robustness to variation that is present in real world settings (Nis Web-Collected; SYNTHIA; SCFACE; COCO-Text) and diverse testing data for providing better estimates of real-world performance (300-W; PPB; MORPH; IIITD). For example, the authors of Nis Web-Collected Database emphasized that diversity of age data could lend to universal age estimation:

\begin{quote}
    \emph{“[T]he derived human age estimator is \textbf{universal owing to the diversity and richness} of Internet images and thus has good generalization capability.”}\,---\,Nis Web-Collected Database
\end{quote}

\begin{quote}
    \emph{``In practice, even the best visual descriptors, class models, feature encoding methods and discriminative machine learning techniques \textbf{are not sufficient to produce reliable classifiers if properly annotated datasets with sufficient diversity are not available}.''} \,---\,SYNTHIA
\end{quote}

Diversity was also motivated as crucial for the effectiveness of deep learning methods, a class of machine learning which methods have a very large number of parameters relative to traditional computer vision methods (NWPU-RESISC45; SYNTHIA). For example:

\begin{quote}
    \emph{``[W]e can generate a \textbf{broad variety} of urban scenarios and situations, which we believe is very \textbf{useful to help modern classifiers based on deep learning}.''} \,---\,SYNTHIA
\end{quote}

\begin{quote}
    \emph{``In addition, almost all existing datasets have a number of limitations, including the small scale of scene classes and the image numbers, the \textbf{lack of image variations and diversity}, and the saturation of accuracy. These limitations \textbf{severely limit the development of new approaches especially deep learning-based methods}.''} \,---\,NWPU-RESISC45
\end{quote}

We found notions of diversity or variety to be closely coupled with the concept of ``natural'' data. For example, creators of 300-W created a dataset of \emph{``naturalistic, unconstrained face images''} that included variations such as \emph{``unseen subjects, pose, expression, illumination, background, occlusion, and image quality.''} This connection was also visible in the way dataset creators described diversity as a property that emerges naturally from ``unconstrained'' data collection methods:

\begin{quote}
    \emph{``The images in this collection display \textbf{large variation in pose, lighting, background and appearance}. Some of these variations in face appearance are due to factors such as motion, occlusions, and facial expressions, which \textbf{are characteristic of the unconstrained setting} for image acquisition.''} \,---\,UMass FDDB
\end{quote}

Generally speaking, we found dataset diversity to be closely coupled with notions of realistic data and challenging data. Diverse datasets were ones that more closely mimicked the real world, and in doing so presented new challenges. 

\subsubsection{Unbiased Data}
Dataset creators frequently motivated unbiased data as desirable\footnote{One notable exception to this is the Animals on the Web dataset that described intentionally inducing a bias between the training and test distributions so that the test set was biased towards more ``difficult'' examples. 
}, often connecting dataset bias to issues of generalization. For example, some attribute a model's failure to generalize\,---\,from one dataset to another, or from a dataset to the real world\,---\,to the existence of dataset bias (KinFaceW; KITTI). In a similar vein to discussions of diversity, unbiased benchmark datasets were motivated through their role in comparing and standardizing algorithms (Imagenet (Paper 2)). Discussions of dataset biases were generally divorced from a broader examination of social or cultural bias or the impacts of dataset bias on individuals from different sociodemographic groups. The only dataset in our corpus to explicitly connect socio-demographically biased data with discriminatory outcomes was PPB:

\begin{quote}
    \emph{``It has recently been shown that \textbf{algorithms trained with biased data have resulted in algorithmic discrimination}...''} \,---\,PPB
\end{quote}

Otherwise, biased data was implicitly understood to be a negative property, and one that might impact classification performance at a more general level. Bias was generally discussed in relation to the process of data collection or properties of images themselves. For example, selection bias, photographer bias, recency bias, and biases in object/person pose or illumination were all topics of discussion (VAIS; PASCAL; COCO-Text). Despite frequent references to unbiased data, we observed that many claims of unbiased or less-biased data remained largely unqualified. For example, some dataset creators describe data that has been created without the particular machine learning task in mind as less biased with no further justification or explanation (PASCAL; COCO-Text):

\begin{quote}
    \emph{``The use of personal photos which were not taken by, or selected by, vision/machine learning researchers \textbf{results in a very 'unbiased' dataset}, in the sense that the photos are not taken with a particular purpose in mind i.e. object recognition research.''} \,---\,PASCAL
\end{quote}

Mentions of bias frequently overlap with discussions of diversity or variety, with the assumption being that sufficient variation across a particular aspect of the dataset minimizes the potential for bias. For example, the creators of INRIA Pedestrian state:

\begin{quote}
    \emph{``Many are bystanders taken from the image backgrounds, so \textbf{there is no particular bias on their pose}.''} \,---\,INRIA Pedestrian
\end{quote}

The desire for unbiased data was intricately tied to controlling for human bias in data collection and annotation. We discuss how dataset authors work to manage human bias in Section \ref{human-bias}.

\subsubsection{High-Quality Data}
\label{high-quality}

Dataset creators frequently described their datasets as high quality. The dataset creators that specified what they viewed to be ``high quality'' data tended to focus on two aspects of datasets: the images themselves and/or the annotations associated with the images. High quality images were generally described as those captured by high quality sensors and as having high resolution (UG\^{}2; NOAA Fisheries; Urban Stereo Scene). For example, the authors of the Urban Stereo Scene Labeling Benchmark Dataset declared the higher resolution of their images in comparison to previous datasets:

\begin{quote}
    \emph{``To our knowledge, the only comparable urban segmentation dataset with stereo vision data has been proposed by [16], which \textbf{our dataset exceeds in terms of ... image resolution} (1024 × 440 px vs. 360 × 288 px ), which is an essential factor for appearance-based segmentation.''} \,---\,Urban Stereo Scene Labeling Benchmark Dataset
\end{quote}

Many of the datasets were also described as having high-quality labels (BigHand2.2M; Stanford Region Labeling; IterNet RGB-D; PASCAL; SUN RGB-D; MS-COCO; ImageNet (Paper 1); Animals on the Web; Middlebury Stereo; BSDS300). High quality labels were often defined in terms of label accuracy relative to pre-specified ``gold standard'' labels or label consistency across different annotators (Stanford Region Labeling; ModaNet; ImageNet (Paper 1)). Other dataset creators used the performance of models trained on the dataset to validate the accuracy of annotations (BigHand2.2M). The authors of the BSDS300 dataset describe how they obtained high quality labels through their annotation interface:

\begin{quote}
    \emph{“In addition to simply splitting segments, the user can transfer pixels between any two existing segments. This provides a tremendous amount of flexibility in the way in which users create and define segments. The interface is simple, yet accommodates a wide range of segmentation styles. In less than 5 minutes, one can create a \textbf{high-quality, pixel-accurate segmentation with 10–20 segments using a standard PC.}”} \,---\,BSDS300
\end{quote}

Datasets containing exclusively high quality images, in terms of clarity and pixel values, were often viewed as oppositional to realistic or challenging data, which we highlight in Sections \ref{diverse-data} and \ref{challenging-data}. For this reason, some dataset creators emphasized the importance of including images with varying levels of quality in order to better reflect the real world:

\begin{quote}
    \emph{``There are \textbf{large variations in the quality of the contributed photographs}, lighting, indoor vs outdoor environments, body shapes and sizes of the people wearing the clothing, depicted pose, camera viewing angle, and a huge amount of occlusion due to layering of items in outfits ... These characteristics reflect the \textbf{extreme challenges and variations that we expect to find for clothing retrieval in real-world applications}.''} \,---\,Exact Street2Shop
\end{quote}

However, high-quality annotations were nearly universally sought after. An overall high-quality dataset was one that was useful and accurately annotated, even if image quality was purposefully varied.

\subsubsection{Realistic Data}
\label{realistic-data}
Realistic data was described as data that was reflective of more ``natural'' conditions, in terms of not controlling for lighting, pose, expression, angle, occlusion, or other image factors. ``Realistic'' was often used to describe data captured in an uncontrolled or unposed settings (UG\^{}2; ND-TWINS-2009-2010; KinFaceW; SFEW; SCFACE); collected from the internet (KinFaceW); collected in a public spaces (SCFACE); or data that is varied along several dimensions such as sensor quality (SCFACE). Some dataset creators described the realism of scene arrangements, for example, by describing a computer mouse on the floor as ``unrealistic'' (SUN RGB-D). Finally, datasets that consisted of synthetic imagery described photorealism as a desirable property that they strived to achieve (IterNet RGB-D; GTA5; SYNTHIA; Middlebury Stereo). 

Dataset creators describe realistic data as being critical for measuring and comparing model performance (George Mason University Kitchen; CUAVE; SCFACE); for developing models that generalize to real-world settings (BiosecurID; Street2Shop; Subway; Middlebury Stereo); and, more generally, advancing research (SFEW). For example, the authors of BiosecurID discussed how data generated through laboratory experiments is of limited utility when estimating model performance in the real world: 

\begin{quote}
    \emph{``In order to overcome the difference in performance between laboratory experiments and practical implementations, there is an \textbf{urgent need for the collection of realistic multimodal biometric data} which permits to infer valid results from controlled experimental conditions to the final application.''} \,---\,BiosecurID
\end{quote}

We observed close ties between the notions of realism and variety, both of which are understood to give rise to more challenging datasets. We discuss these properties in more depth in the next section, including how realistic data is challenging.

\subsubsection{Challenging Data}
\label{challenging-data}

Dataset difficulty is another characteristic often touted by dataset creators. Dataset creators frequently motivate the creation of a new dataset with reference to saturated model performance on previous, easier datasets. For example, the creators of INRIA Pedestrian state that their new algorithm gives \emph{``essentially perfect results on the MIT pedestrian test set, so we have created a more challenging set''} (INRIA Pedestrian). Dataset creators also describe the importance of new and more challenging datasets for advancing progress in the field (SCFACE; DUT-OMRON; ImageNet (Paper 1)) and mitigating the potential for methods to overfit (LFW). For example:

\begin{quote}
    \emph{“As computer vision research advances, larger and \textbf{more challenging datasets are needed} for the next generation of algorithms.”} -- ImageNet (Paper 1)
\end{quote}

The characteristics that constitute challenging data tend to relate to variability along a variety of dimensions including, but not limited to, sensor quality, image scale, illumination, object or person pose, camera viewpoints, and locations of data collection (INRIA Pedestrian; Stanford Region Labeling; ND-QO-Flip; OSU Thermal; Exact Street2Shop; Leeds Sports Pose; H3D; VAIS). For example, the authors of the OSU Thermal Pedestrian Database wrote that their data was challenging because of varying backgrounds and thermal intensities, and they want to collect more challenging data for future work:

\begin{quote}
    \emph{``The approach was demonstrated with a \textbf{difficult dataset of thermal imagery with widely-varying background and person intensities}. In future work, we plan on \textbf{extending the dataset to include additional situations involving many more distractors} moving through the scene.''} \,---\,OSU Thermal Pedestrian Database
\end{quote}

As touched on in Section \ref{realistic-data}, challenging data was often characterized as data that mimics real-world (unconstrained) conditions and real-world variation:

\begin{quote}
    \emph{``Similar to conventional face recognition, when the targets are unconstrained faces in the wild [12] (i.e., variation in pose, illumination, expression, and scene) the \textbf{difficulty level further increases}, and the same being true for kinship recognition. These are, unfortunately, challenges that need to be overcome. Thus, FIW \textbf{poses realistic challenges needed to be addressed before deploying to real-world applications}.''} \,---\,FIW
\end{quote}

In short, challenging data was compatible with realistic and diverse data, but often incompatible with high-quality data, in terms of resolution quality. Low resolution images proved more challenging than high resolution images.

\subsubsection{Comprehensive and Large-Scale Data}
\label{big-data}

Dataset creators often described their datasets as large-scale, indicating their datasets contain a large number of data instances and, in some cases, a large number of categories. The importance of large-scale datasets, or ``big data,'' are varied. Some creators emphasized the importance of large datasets for improving generalization performance and reducing risks of overfitting (BigHand2.2M; LFW; SUN RGB-D). Relatedly, some creators identified large datasets as key to the development of reliable models and reliable evaluation methods (OU-ISIR Gait). But a number of dataset creators describe large scale datasets as being broadly essential to advancing computer vision research (KAIST; ImageNet (Paper 1 \& 2); BigHand2.2M). Related to the above category of standardization and benchmarking, dataset creators specifically called out the importance of large-scale benchmark datasets:

\begin{quote}
    \emph{``Because data-driven AI-based methods have enabled breakthroughs in both academia and industry, \textbf{large-scale benchmarks have become one of the most important factors} to advance this technology.''} \,---\,KAIST
\end{quote}

Several authors specifically identified the importance of big data for advancing deep learning methods in particular (NWPU-RESISC45; SYNTHIA; BigHand2.2M; One-Million Hands; GTA5):

\begin{quote}
    \emph{``However, the \textbf{lack of publicly available `big data' of remote sensing images severely limits} the development of new approaches especially deep learning based methods.''} \,---\,NWPU-RESISC45
\end{quote}

Many dataset creators included claims of dataset completeness or comprehensiveness, often referencing the  dataset size or high variability of data instances (FIW; TUM GAID; CUAVE; BigHand2.2M). Others describe the categories or annotations that structure the dataset to be comprehensive or complete (GTA5; VOC; SUN; ImageNet (Paper 2)). The ImageNet creators claim their dataset provides the \emph{``most comprehensive and diverse coverage of the image world''} (ImageNet (Paper 1)). Similarly, the authors of SUN wrote that their dataset of scenes contained all discursively important images:

\begin{quote}
    \emph{“First, we seek to quasi-exhaustively determine the number of different scene categories with different functionalities. Rather than collect all scenes that humans experience - many of which are accidental views such as the corner of an office or edge of a door - \textbf{we identify all the scenes and places that are important enough to have unique identities in discourse, and build the most complete dataset of scene image categories to date}.”} \,---\,SUN
\end{quote}

Large-scale and comprehensive data was compatible with notions of challenging and realistic data. As described further in Section \ref{annotation-labor}, high-quality annotations become increasingly challenging and costly as datasets grow in size and scope.

\subsection{Human Actors as Annotators and Data Subjects}

Authors often discussed the presence of human actors in the dataset construction and documentation process, ranging from the authors themselves to annotators and data subjects. By and large, human considerations were often distinctly aimed at making the data collection and annotation processes objective. By objective, we mean that there was an attempt to mitigate or remove human subjectivity from all data processes. This was particularly salient when discussing the role of data collection and annotation. Human subjectivity was seen as a detriment to consistent, ``clean,'' comprehensive, and accurate data and annotations\,---\,the properties of desirable data. Less commonly, dataset authors also discussed the diversity of human data subjects, from the perspective of the technical benefit of diverse human subjects for developing models. In this section, we discuss the four different human considerations we found: \emph{annotation labor and time costs; human properties as a barrier; humans as diverse data;} and \emph{human bias in data collection and annotation}.

\subsubsection{Annotation Labor and Time Costs}
\label{annotation-labor}

A number of datasets utilized human annotation (63 of 114; 55\%). Within papers, there is a variable amount of information available on the identity of human annotators and the cost of their labor. In cases in which human annotation is used, human annotators are described in 40 of 63 (63.4\%) cases. Of these, 23 of 63 (36.5\%) were reported as third-party workers (mostly from Amazon Mechanical Turk); 9 of were the authors; 6 were students; and 5 were some mixture of these. Only 5 of 63 (7.8\%) papers report annotator demographics, and only 4 of 63 (6.3\%) papers report if annotators were compensated.

A major focus in discussing human annotation was the time and monetary cost of annotation, particularly as a barrier to annotating large-scale datasets (BigHand2.2M; Leeds Sports Pose; UNBC-McMaster; Stanford Region Labeling; ModaNet; SYNTHIA; MS-COCO; PASCAL; ImageNet (Paper 1); LFW; GTA5), which were otherwise highly valued by dataset authors, as highlighted in Section \ref{big-data}. High costs negatively result in \emph{``slowing down the development of new large-scale collections like ImageNet''} (SYNTHIA). Yet, manual annotation was also seen as desirable for ``high-quality'' data. For example, the authors of SYNTHIA wrote of the necessity but difficulty of having large amounts of annotated data:

\begin{quote}
    \emph{``Having a sufficient amount of diverse images with class annotations is needed. These annotations are obtained \textbf{via cumbersome, human labour} which is particularly challenging for semantic segmentation since pixel-level annotations are required.''} \,---\,SYNTHIA
\end{quote}

Here, we can see the SYNTHIA authors express the need for human labor. This need stemmed from the view that manual labelling was more accurate than automated labelling, and ground truth accuracy of labels is a desirable data property (see Section \ref{high-quality}). However, while manual labelling was often prized for its accuracy, authors tried to minimize the amount of time and money spent on human annotation labor (BSDS300; FIW; ImageNet (Paper 1); MS-COCO; Stanford Region Labeling). Often, Amazon Mechanical Turk and other crowdworking platforms were viewed as an extremely valuable tool for the computer vision community, particularly due to the demand for a great deal of labor at a low cost (SUN RGB-D; SUN; MS-COCO; ImageNet (Paper 1); H3D; Stanford Region Labeling; CORE). The following examples showcase authors touting minimal labor and low costs:

\begin{quote}
    \emph{``We now discuss the procedure followed to collect, organize, and label 11,193 family photos of 1,000 families with minimal manual labor.''} \,---\,Families in the Wild (FIW)
\end{quote}

\begin{quote}
    \emph{``We constructed using Amazon Mechanical Turk (AMT), at a total cost of less than \$250.''} \,---\,Stanford Region Labeling Dataset
\end{quote}

The use of human annotators and explication of human annotations as valuable to computer vision signals the importance of human beings in the modeling pipeline. Yet, there is also the goal of minimizing human labor costs, suggesting a devaluing of labor that is otherwise valuable to the process of dataset curation. Further, there is also an underlying expectation that human annotators are flawed, in that they make mistakes or introduce subjective beliefs that do not align with the expectations of the authors.

\subsubsection{Human Properties as a Technical Barrier}

Much like the subjectivity of human annotation can risk the desired objectivity of labeling, human behaviors may be seen as technically problematic to either building a dataset or the resulting accuracy of a model. The complexity of human properties was generally perceived as a barrier to human data collection and task specification. Human characteristics were portrayed as difficult to control ranged from the diversity of human appearance (UMass FDDB; SHEFFIELD; ND-QO-Flip; MORPH; BP4D-Spontaneous; CMU-MultiPIE; ModaNet; Paper Doll; INRIA Pedestrian; TUM GAID; Leeds Sports Pose; OU-ISIR Gait; Exact Street2Shop) to the autonomy of human decision making (M3; IIITD ). For example, the authors of Leeds Sports Pose described the difficulty of pose estimation given the way range of human appearance and natural imaging conditions:

\begin{quote}
    \emph{``The task is particularly challenging \textbf{because of the wide variation in human appearance present in natural images due to pose, clothing and imaging conditions}.''} \,---\,Leeds Sports Pose
\end{quote}

Notably, no authors discuss ethical considerations in their use of the data. Human autonomy posed issues to accessing data deemed necessary for the desired task. For instance, the authors of M3 discussed how it was difficult to collect biometric data, like fingerprints, due to subjects revoking their consent due to privacy concerns. Human desires for privacy makes some data difficult to come by:

\begin{quote}
    \emph{``We found that collecting fingerprint data is especially difficult because some recruited subjects later \emph{decided that they are reluctant to provide their fingerprint data due to privacy concerns.}''} \,---\,M3
\end{quote}

Similarly, the authors of IIITD Plastic Surgery Face Database described having difficulty building a dataset of before-and-after plastic surgery images. They discussed that people did not wish to share these images with them or online due to privacy concerns, which they instead scraped from the web as a result: 

\begin{quote}
    \emph{``Due to the sensitive nature of the process and the privacy issues involved, it is extremely difficult to prepare a face database that contains images before and after surgery.''} \,---\,IIITD Plastic Surgery Face Database
\end{quote}

We noted that even datasets with especially sensitive human data, such as the nude detection dataset by Lopes et al., where nude images were collected from the web, had no discussion of ethics, privacy, or even an ethics review process. Only 5 of the 100 datasets (5\%) containing human subjects mentioned having an IRB or international equivalent (Forensic Facial Examiner Study; HumanEva; MORPH; ND-TWINS-2009-2010; CAFE). Five mentioned privacy considerations in any capacity (Beauty 799; BiosecurID; FACEBOOK100; KITTI; SCFACE). For example, the authors of SCFACE outlined limiting access to specific subject images due to privacy concerns:

\begin{quote}
    \emph{``For legal reasons and for the privacy of the database participants, images that can appear in reports, papers, and other documents published or released are those with subjectID: 001, 002, 045 or 102 in the SCface database.''}  \,---\,SCFACE
\end{quote}

When performance failures occurred, such failures were sometimes attributed to issues of human diversity or behaviors. Such failures might occur when training data is too controlled to reflect the real world. For example, the authors of OU-ISIR Gait Database wrote of their model’s gait classification failures:

\begin{quote}
    \emph{``These failures mainly originated from the unique walking style (e.g., some subjects raise their arms higher than generic subjects) or special clothing (e.g., a long dress or coat), which cause a large difference between this test sample and the generic training samples.''} \,---\,OU-ISIR Gait Database, Large Population Dataset with Age
\end{quote}

Given the barriers presented by human-based data, whether face or body, some authors would describe how they mitigated or bypassed such barriers. For example, authors might trade-off real-world diversity of appearance by attempting to implement controlled conditions (inside or outside of studio settings) during the data collection process (e.g., M3). In the case of privacy or consent concerns, authors have instead scraped images from the web to surpass participant autonomy (e.g., IIITD).

\subsubsection{Humans as Diverse Data}

As previously highlighted in Section \ref{diverse-data}, diversity took on a very specific meaning: describing diverse instances of data. This most commonly included such instances as a diversity of object types, a diversity of lighting conditions, and a diversity of angles. The term ``diversity'' was not commonly used in terms of human conditions, such as race, gender, or ability. Even in cases where the task was tied to human conditions, like age, race, or gender, authors did not necessarily discuss diversity in terms of representation. Of the datasets containing images of humans, 41 of 100 (41\%) provided information about the sociodemographic diversity of data subjects.

When used to describe humans, diversity was most often attributed to diversity of ages (CASIA NIR-VIS 2.0; FIW; Nis Web-Collected; OU-ISIR Gait; RAFD; UvA-NEMO Smile; MORPH; SFEW). Other occurrences included diversity of race or ethnicity (PPB; CAFE; MORPH; Ethnic DB; FIW). Diversity of gender was rarely discussed in our sample; gender was entirely binary, as previous research has found \cite{Scheuerman2020}, and few datasets discussed the distribution of men versus women (PPB). Diversity statements were often written like those in CAFE and The CASIA NIR-VIS Database, describing that diversity in the data as a feature:

\begin{quote}
    \emph{``It is also \textbf{racially and ethnically diverse}, featuring Caucasian, African American, Asian, Latino (Hispanic), and South Asian (Indian/Bangladeshi/Pakistani) children.''} \,---\,The Child Affective Facial Expression (CAFE)
\end{quote}

\begin{quote}
    \emph{``In the new database, \textbf{the age distribution of the subjects are broader}, spanning from children to old people.''} \,---\,The CASIA NIR-VIS 2.0 Face Database
\end{quote}

Much like claims about diverse data instances in Section \ref{big-data}, there were attempts to claim universality of the human diversity captured in the data. The authors of Nis Web-Collected Database claimed a universality in age estimation due to the diversity of their data:

\begin{quote}
    \emph{``The derived human age estimator is \textbf{universal owing to the diversity and richness of Internet images} and thus has good generalization capability.''} \,---\,Nis Web-Collected Database
\end{quote}

Generally, diversity was posited as a technical benefit. That is, a diversity of human characteristics benefits the technical accuracy of the model proposed. A few dataset authors discussed benefits beyond the technical; that is, given the potential deployment of the system itself, how it could socially benefit people to use diverse human data. For example, the authors of PPB discuss how groups underrepresented in data would suffer from the resulting lower accuracy rates but nonetheless suffer social consequences:

\begin{quote}
    \emph{``In other contexts, a demographic group that is \textbf{underrepresented in benchmark datasets can nonetheless be subjected to frequent targeting}.''} \,---\,PPB
\end{quote}

However, diversity of the humans involved in the overall creation of the dataset was attributed primarily to data subjects. Most authors did not discuss the diversity of annotators or those involved in the data collection process. There were a few exceptions. Five datasets included demographic information on annotators (Beauty 799; Forensic Facial Examiner Study; ModaNet; RAFD; CAFE). Of these 5 datasets, most focused on gender and age distribution, but 2 also provided ethnicity information (CAFE; Beauty 799). For example, the authors of CAFE described the demographic distribution of their annotators:

\begin{quote}
    \emph{``One hundred undergraduate students (half male, half female) from the Rutgers University-Newark campus participated (M = 21.2 years) … The sample was 17\% African American, 27\% Asian, 30\% White, and 17\% Latino (the remaining 9\% chose ‘Other’ or did not indicate their race/ethnicity).''} \,---\,The Child Affective Facial Expression (CAFE)
\end{quote}

Diversity in human data stood to ensure data users of the utility of the data, in terms of being unbiased and ideally more accurate for all groups. Though not explicitly described, the few instances of described diversity in annotator demographics insinuated to data users that representation in annotation would result in less biased or skewed labels.

\subsubsection{Human Bias in Data Collection and Annotation}
\label{human-bias}

As demonstrated throughout this section, much discussion about the role of human beings in constructing and annotating datasets is around controlling for human behaviors. Another aspect of controlling human behaviors is mitigating human bias. When discussing the roles of researchers and annotators, it was often to ensure readers that human bias was accounted for and mitigated in the collection (Multi-Spectral Pedestrian; Abstract Paintings / Artistic Photographs; COCO-Text; UG\^{}2; VAIS; Animals on the Web; PASCAL) and/or annotation process (BigHand2.2M; PETS04; SYNTHIA; ImageNet (Paper 1)). For example, the authors of the PASCAL VOC dataset wrote that their data collection process resulted in unbiased images:

\begin{quote}
    \emph{``The use of personal photos which were not taken by, or selected by, vision/machine learning researchers \textbf{results in a very `unbiased' dataset, in the sense that the photos are not taken with a particular purpose in mind} i.e. object recognition research.''} \,---\,PASCAL VOC
\end{quote}

Some authors would justify their choice of collecting certain types of data due to potential biases. For example, the authors of the VAIS dataset decided to collect images to avoid the ``photographer bias'' they associate with web images. The authors of COCO-Text discussed how the MS COCO dataset’s collection method makes it a good source for unbiased text images:

\begin{quote}
    \emph{``\textbf{Images from the web often suffer from photographer bias}, in that images with more aesthetic appeal tend to be uploaded.''} \,---\,Maritime Imagery in the Visible and Infrared Spectrums (VAIS)
\end{quote}

There was also concern that, although manual annotation was viewed as valuable (One-Million Hands; Texas 3D; CUAVE; UAVDT; USF; PETS04), and sometimes are superior to automated annotation (ModaNet; SUN), that human subjectivity would result in inaccuracies (BigHand2.2M; PETS04; SYNTHIA; ImageNet (Paper 2)). For example,  the authors of ImageNet (Paper 2) explained two sources of bias that they attempted to control for during the annotation process:

\begin{quote}
    \emph{``While users are instructed to make accurate judgment, \textbf{we need to set up a quality control system to ensure this accuracy}. There are two issues to consider. First, human users make mistakes and not all users follow the instructions. Second, users do not always agree with each other, especially for more subtle or confusing synsets, typically at the deeper levels of the tree.''} \,---\,ImageNet (Paper 2)
\end{quote}

Given the risk of human subjectivity and its associated errors, many authors employed checks to mitigate human subjectivity (PASCAL; 300-W; ImageNet (Paper 2)). Therefore, they could benefit from the robustness and accuracy of human visual assessments in their annotation, while ensuring those annotations met their expectations. For example, the authors of the PASCAL VOC challenge and its associated dataset discuss how the organizers of the challenge employed manual checks on annotations to ensure they were correct, in terms of what was expected:

\begin{quote}
    \emph{``Following the annotation party, \textbf{the accuracy of each annotation was checked by one of the organisers}, including checking for omitted objects to ensure exhaustive labelling.''} \,---\,PASCAL VOC
\end{quote}

Through these statements, dataset creators meant to ensure potential users of their creations that author and annotator contributions would be unbiased. 

\section{Discussion}

The increasing application of computer vision technologies in public life presents profound stakes for how human beings interact with the world. The tasks that computer vision models are designed to do are ancillary to the data used to train, test, and validate those models. Data is crucial to the process of model design, and therefore to the advancement of the field of computer vision. Thus, it is a key point of the computer vision pipeline for examining how values become embedded into the technical artifacts designed by researchers and practitioners in the field. 

In this paper, we conducted a large-scale analysis, focusing on a range of disciplinary practices in dataset collection, curation, annotation, and release, across both human and non-human related tasks. Our analysis centered around common themes previously identified in prior literature on machine learning datasets, including data annotation, data availability, data categories, and data collection processes. Through a structured and qualitative content analysis, we uncovered both explicit statements and silences about data, and what those statements and silences implied about the values of datasets in computer vision. Overall, we found that computer vision dataset authors valued \emph{efficiency, universality, impartiality,} and \emph{model work}. 

We characterize the values of computer vision datasets by discussing them as trade-offs to otherwise silenced values: \emph{efficiency versus care, universality versus contextuality, impartiality versus positionality,} and \emph{model work versus data work}. The values explicit in computer vision datasets often ignored, or implicitly critiqued, values of human-centered computing approaches. For example, social computing has embraced value-centered and value-sensitive design \cite{friedmanValuesensitiveDesign1996} that includes considering context \cite{broensCapturingContextRequirements2007, taylorSituationalWhenDesigning2017, chancellorWhoHumanHumanCentered2019}, reflexivity and positionality (e.g., \cite{Garcia2019, Kaeser-Chen2020}), and situated expertise (e.g., \cite{MacKay1999, Easley2018, Kempe-Cook2019}). For each silence, we offer recommendations for dataset authors to begin to address values in the design process, drawing on concepts from prior work in social computing and algorithmic fairness. 

Many recommendations are aimed at actions that individual dataset authors can take, but we acknowledge the role of larger institutional incentive structures that may prevent individuals from effectively implementing change. Thus, we would similarly encourage larger institutions \,---\, conference venues, journals, and academic departments \,---\, to engage with our recommendations at an institutional level. For instance, NeurIPS has recently developed a ``Datasets and Benchmarks Track''\footnote{https://neuripsconf.medium.com/announcing-the-neurips-2021-datasets-and-benchmarks-track-644e27c1e66c} to incentivize work on machine learning datasets and as ``an incubator to bootstrap publication on data and benchmarks.'' Further, while we focus specifically on computer vision given the empirical focus of this paper, the highlighted values and silences, and subsequent recommendations, can be abstracted to other machine learning domains \,---\, though we would still advocate specific values analyses for each domain.

\subsection{Efficiency over Care}
\label{sec:efficiency}

Dataset authors valued efficiency, both in terms of the time spent and the associated costs, monetarily and computationally, of gathering and annotating data. Authors sought desirable properties, in terms of objective, unbiased, neutral, and comprehensive data, that are easily available, quickly and cheaply classifiable, and able to be quickly but accurately annotated. The value of efficiency was clear in a number of practices in our findings. In terms of higher-level disciplinary practices, a focus on efficiency may have led many dataset authors to document only the barest technical details of the dataset creation process. Few authors wrote more than a few paragraphs documenting their dataset practices, insinuating not only a focus on model work over data work (see Section \ref{model-work}), but an efficient and highly condensed means of conveying the technical details viewed as most important to the dataset. In seeking desirable data, many authors employed the practice of scraping publicly available data from websites, seen as both an easy and cheap method of amassing large amounts of data in a short period of time. Human concerns about privacy and data ownership were sometimes posited as barriers to collecting data, while the reasonings for those concerns were otherwise ignored.

Some dataset creators invested time, labor, and money into dataset debiasing and quality control measures in the service of high quality data and annotations. These investments were often justified with respect to the potential gains in algorithmic advances that stem from large-scale datasets with quality annotations. These investments were often framed as a necessary cost, but still a cost to be minimized as much as possible. Many authors employed crowdworkers for annotating that data, explicitly due to the low cost of paying crowdworkers for annotation work. Further, crowdsourcing platforms like Amazon Mechanical Turk allowed authors to hire many cheap laborers at once, leading to faster annotations. Untrained annotators were often framed as desirable, with some dataset developers investing significant efforts in developing their own annotation interfaces that would allow untrained annotators to be utilized en masse (e.g., ScanNet).

The lack of attention paid to the perspectives, labor, and rights of human actors in the dataset curation process points to the lack of value in the field in explicating or thinking through the human role in technical practices. As discussed by \citeauthor{chancellorWhoHumanHumanCentered2019}, the discursive practices in defining the human subject in machine learning is both dehumanizing and a threat to scientific rigor \cite{chancellorWhoHumanHumanCentered2019}. This is particularly salient in datasets with highly sensitive data, like the referenced nudity detection dataset by Lopes et al., which proposed nudity detection as a form of object detection but did not engage with social issues of privacy or gender-based bias in their documentation. 

Valuing efficiency was at the cost of care, valuing slow and thoughtful decision-making and data processes, considering more ethical ways to collect data and treat annotators, and seeking fairer compensation\,---\,or even reporting compensation\,---\,for data labor. Generally, compensation for either data or annotation labor was not reported. Further, authors did not discuss the costs of efficiency to ethical scientific practice or potential harmful social implications, such as publishing sensitive data (e.g., Lopes et al.’s nudity dataset) or contributing to the documented class divisions between technology experts in industry and academia and the gig economy associated with crowdworkers \cite{fort2011amazon, pittmanAmazonMechanicalTurk2016, williamson_2016}. In general, there was little to no discussion about ethics when conducting work with annotators or with human subjects as data instances. Even commonly accepted forms of ethics accountability were not regularly reported by dataset authors, such as IRB or international equivalents of institutional ethics checks. 

A counterexample within our corpus which prioritized care within the process was the Child Affective Facial Expression (CAFE). The authors of CAFE sought explicit consent from parents to collect face data from their children. The dataset collection went through IRB review, who deemed the children a vulnerable population and required the authors to outline potential harms and expected benefits to participants. While data collected in a studio setting and required participants to explicitly consent may result in a far slower and more laborious data collection process, it also shows mindfulness towards the privacy and autonomy of human data subjects. To further incorporate a value of care into their processes, the authors of CAFE might have included ethical considerations of data use and decided to compensate both data subjects and annotators. 

\subsubsection{Recommendations for Incorporating the Value of Care}

Valuing care might lead to less efficient, but more reflexive data practices \cite{miceliDocumentingComputerVision2021}. While not an exhaustive list of potential approaches, we highlight several areas for dataset authors to embrace care over efficiency:

\begin{itemize}
    \item Collect data with special attention to the privacy and ownership rights of those data. This may mean going through the laborious step of obtaining permission from copyright holders (for example, when scraping Flickr for object recognition tasks) or data subjects themselves (for example, when collecting images of human faces). Given the significant overhead of studio data collection processes, and its associated deficiencies in terms of realism and diversity, authors should consider reaching out to copyright holders on social media sites for explicit permission of data use, instead of otherwise scraping those photos without permission. Asking for explicit use could allow a balance between efficiency and care, and still allow dataset authors to collect properties of desirable data. Initiatives around data licenses with differing data use permissions (e.g., \cite{contractorResponsibleAILicenses}) are useful resources for streamlining collecting data more ethically, and initiatives around labels offer appropriate labels for specific groups (e.g., for indigenous groups \cite{anderson2013chuck}.
    
    \item Compensate data subjects for their data and annotators for their labor. Many institutions, in both academia and industry, have fair contribution guidelines for human subjects research. \citeauthor{litmanRelationshipMotivationMonetary2015} found that high-quality responses on Amazon Mechanical Turk necessitated payments above the minimum wage \cite{litmanRelationshipMotivationMonetary2015}. Dataset authors should consider the context of compensation, in terms of what is most valued by annotators; money or gift cards may not be the most appropriate compensation for all contexts. For example, \citeauthor{hodgeRelationalFlexibleEveryday2020} found that money was not useful to their participants who live in care homes \cite{hodgeRelationalFlexibleEveryday2020}. Further, if the authors expect to make a profit from their dataset, they should consider what fair compensation would be to their subjects and authors given that profit necessitates their data and labor. 
    
    \item Institutional review board (IRB) or ethics reviews, while certainly imperfect, are not the norm in machine learning; often, because they are not required for scraping data, researchers do not submit an IRB at all \cite{metcalfWhereAreHuman2016}. Normalizing IRB and ethics review processes for machine learning is an ongoing conversation within machine learning communities \footnote{For example, a recent workshop on "Navigating the Broader Impacts of AI Research" was hosted as part of the top-tier machine learning conference NeurIPS (https://nbiair.com/)}. However, IRBs and other institutional ethics review processes could provide checks to minimize harm to data subjects or annotators during the dataset curation process. Given the nature of machine learning data to have downstream impacts on the model, researchers should also report predicted uses of the data and how it may implicate the privacy of human data subjects.
    
    \item Store and safeguard datasets with proper data stewardship protocols in place, such as gatekeeping access to data, terms of service and potential licensing agreements. Institutional repositories such as the ICPSR and university libraries can aid in this process and implement best practices for data curation and preservation \cite{inter-universityconsortiumforpoliticalandsocialresearchGuideSocialScience2012}. Dataset authors can work with librarians and data stewards to provide open access to other researchers with proper agreements which set conditions and terms of use. Dataset authors might look to initiatives such as IBM's data privacy passports to protect sensitive data across multiple cloud infrastructures \cite{ibmIBMDataPrivacy}. An added benefit is that data repositories can record histories of data use and access, which can open up the ability of external researchers to independently audit these histories. 
\end{itemize}

\subsection{Universality over Contextuality}

We found computer vision dataset creators valued large-scale, diverse, and realistic data that lent to a belief in inherently comprehensive or complete categorical classifications of real-world phenomena. We observed the widespread valuation of these properties was rooted, in large part, in an assumption that larger and more varied datasets provide better approximations of the real world and thus afford the development of high capacity models that are able to generalize well to varied real-world settings. Implicit in this belief is the value of universality, insinuating a world that is able to be neatly captured and classified, often for the purposes of state and economic management \cite{scottSeeingStateHow2008, murphyEconomizationLife2017, koopmanHowWeBecame2019}. We also observed annotation practices were frequently portrayed as objective; annotation quality checks were aimed at ensuring annotators' worldviews matched the dataset author’s. Universal data properties were viewed as valuable not only to real-world application, but also to standardization and reproducibility. It would be difficult if not impossible to standardize classifications about the world. Framing quality around a presumed objectivity of labels suggests a universal ordering. 

Despite diversity posing more difficulty to fully capturing the world, dataset author's implicitly acknowledged that the world is full of diverse data. Capturing that diverse data in the dataset meant that the model’s observation of the world would be more complete. This included not only a diversity of different types of objects for object recognition tasks, but a diverse array of human beings. For example, the importance of the inclusion of varying lighting conditions and poses was a frequent point of discussion. Authors of human-centered (face- and body-based) datasets also highlighted diverse ages, ethnic categories, and gender distributions, as these were seen as important categories of diversity for human subjects.

Universality was embraced at the expense of \emph{contextuality}, how circumstances such as time, location, or use shape the world and thus the data in a dataset. For example, the geographic origins of images within object recognition datasets were rarely discussed. The language employed to classify objects or people in datasets was not attended to. Why specific identity markers were chosen for representing diversity was absent. Further, how important technical components of explaining the diversity of data, such as lighting, differently affected different groups\,---\,such as those of different ethnic origins\,---\,was never discussed. Beyond categorizing the data itself, authors also rarely discussed the potential impacts of the dataset, and resulting computer vision systems developed from it, on members of different social groups. Datasets were often posited as generalizably useful to broad tasks, such as general object recognition or human detection. A notable exception is the PPB dataset, which was motivated by differential classification scores in computer vision for women with darker skin tones. 

As discussed by prior work, the natural world is actually difficult to capture and classify, and an attempt to reduce socially-shaped categories down into data is difficult \cite{mager-etal-2018-challenges, Vries_2019_CVPR_Workshops, Scheuerman2020}. To embrace the value of contextuality would be to embrace this difficulty and instead focus on the circumstances of use. For example, where data will be used, who should be included in the dataset based on its intended use, and how factors like culture, language, and location should shape the data collected. Further, contextual decisions would need to be motivated by which data subjects might be impacted by data curation decisions and how. LFW provides a good example of contextuality by stating explicitly that there is no naturally accepted distribution of faces for all possible domains. This move acknowledges different worldviews on human diversity. Similarly, NOAA Fisheries was constrained to a very specific context-of-use\,---\,classification to aid in live fish recognition within fisheries\,---\,which informed what kind of data would be collected and how it should be classified. 

\subsubsection{Recommendations for Incorporating the Value of Contextuality}

Context can be scoped in multiple ways---from the context the dataset is meant to be used in (e.g. a specific discipline, work practice or area of application) to the context the dataset is meant to capture (e.g., the diversity of plants in North America). Designing for specific contexts, in terms of specific workplaces, communities, and cultures, has a rich history in HCI and user-centered design that dataset authors could use to guide their work \cite{grudinComputersupportedCooperativeWork1994, clemmensenOverviewDecadeJournal2010, hayesRelationshipActionResearch2011}.

\begin{itemize}
    \item Design datasets for specific temporal, cultural, geographic, or community contexts, rather than for generalized, universal use. Datasets scoped to specific contexts would be more rich and robust for those contexts (e.g., having a dataset of only fish increased the opportunity to robustly capture and classify fish in a useful way than a dataset trying to capture every animal on earth). Similarly, datasets scoped to certain cultures should reflect cultural language and expectations, a method for reducing cultural bias that stems from universalism. 
    
    \item Conduct empirical studies to understand the context of intended use \emph{before} designing the dataset (or associated methods or models). Employing empirical methods, such as surveys, interviews, or ethnography, can ground dataset curation decisions and make datasets more contextually useful for intended stakeholders.
\end{itemize}

\subsection{Impartiality over Positionality}

Although universality was highly valued in our findings, we found that the universe of data was often highly constrained to a specific, impartial worldview. Dataset authors strive for \emph{impartiality} on behalf of the data collectors (often the authors themselves) and the annotators. They seek a debiasing of human subjectivity, so that data is “unbiased” and, implicitly, more trustworthy. Bias in this case takes on a statistical and cognitive definition. In particular, selection biases or observer biases were of major concern to authors. They strived to ensure potential data users that there were no selection biases driving them to select specific types of data (e.g., only images which held aesthetic appeal may have been uploaded to the web for Maritime Imagery in the Visible and Infrared Spectrums (VAIS)). They also tried to mitigate potential observer biases in annotation by focusing on how the subjective perceptions that annotators have would detrimentally lead to inaccurate labelling. 

While there are explicitly stated concerns about introducing human bias, in both the process of data collection and the process of annotation, there is little discussion about how bias, in terms of hermeneutic perspectives or as a matter of interpretation, is unavoidable in vision-based tasks. Dataset authors did not report on their own \emph{positionality}, such as how one's social and professional position can give rise to differential resources and knowledge gaps. For example, industry might afford higher budgets for recruiting data subjects, but individuals may lack the domain-specific information of professionals working in, for example, marine biology (relevant to fishery datasets); nor did dataset authors report how identity characteristics might impact the perspectives of annotators, such as how local or regional culture might influence perspectives on beauty. Instead, they assumed that there are inherently neutral practices to strive for, disregarding the rich scholarly history discussing how all human decisions are inherently value-laden \cite{teoEpistemologicalViolence2014, wallace2019view, zhangVisualCulturalBiases2020}.

Others have argued that this false objectivity lends to a decrease in utility of computer vision data (e.g., \cite{Scheuerman2020}). While dataset authors discuss their work from the perspective of increasing trust and reliability in its objectiveness and utility, they unknowingly decrease that trust and reliability by refusing to describe its most relevant stakeholders. The tasks that datasets are meant to assist both computer vision researchers and domain experts with are inherently human-centered, but datasets lack a human-centered approach to their construction and documentation.

Despite the critical role annotators play in determining the contents of a dataset, dataset creators tended to omit critical details regarding who was performing the annotation tasks. For example, only five publications provided any demographic information regarding who annotated the data instances. As Elizabeth Anderson argues, objections to value-laden inquiry misunderstand the goals and methods of science \cite{anderson1995knowledge}. Values are inherent in all science, and adopting specific subject positions do not diminish scientific practices of reliability and empirics. To better incorporate values of positionality into datasets, authors would need to acknowledge professional and personal identity characteristics of both authors and annotators, and embrace a position that there is no “view from nowhere” \cite{haraway1988situated}. 

A positive example from our analysis includes the Pilot Parliaments Benchmark (PPB) dataset. The authors include rationales for data selection, as well as acknowledgments of the identity-based limitations of their work, specifically around binary gender categories. They also report that their ground truth labels come from a board-certified surgical dermatologist, showcasing how an annotator’s experience and professional training can provide trust, rather than distrust, in the data process.

\subsubsection{Recommendations for Value of Positionality}

\citeauthor{attiaBeComIng2017} provide guidance towards becoming a reflexive researcher, building trustworthiness in a research approach through the expression of personal values and professional skills \cite{attiaBeComIng2017}. We suggest data authors' embrace movements towards reflexivity, rather than attempting to remain impartial and rejecting their own subjectivity or the subjectivity of those with whom they work, such as annotators, collaborators, and other stakeholders.

\begin{itemize}
    \item Positionality statements are one method dataset authors might consider when incorporating reflexive thinking into their work \cite{roseSituatingKnowledgesPositionality1997}. These statements have become accepted practice, not only in feminist scholarship where the concept proliferated, but increasingly in social computing (e.g. \cite{dymComingOutOkay2019, roldanOpportunitiesChallengesInvolving2020, simpsonYouYouEveryday2021}). Positionality statements include researcher reflections on how their own perspectives and values shape the work. For example, these statements can include how one’s disciplinary training lends expertise in certain ways of approaching a research problem, while acknowledging that training's own limitations in other approaches. They also can include how one’s nationality, race, gender, class, or other socio-historical identity impact the context and outcome of a project. Dataset authors should be careful not to incorporate an inherent distrust of the humans helping to shape datasets solely because they have an identity and a perspective. At the same time, we acknowledge that authors should not feel the need to disclose sensitive attributes about their identities that might otherwise endanger them. 
    
    \item Reporting author and annotator demographics may be one useful tool for making transparent how decisions may have been made during the data process. There are numerous guides for collecting and reporting on identity data in ethical and sensitive ways (e.g., \cite{scheuerman2020hci} for gender, \cite{forward2015race} for race). Authors might even consider targeted recruitment of annotators with specific expertises or identity experiences who would be best-suited to annotating the data. For instance, \citeauthor{patton2019annotating} find that annotators who are familiar with gang-related activity have significantly different annotations of Twitter activity than those who are not \cite{patton2019annotating}.
    
    \item Writing ethical considerations is becoming increasingly required for machine learning venues. Authors are being tasked with actively engaging with the potential social and political implications of machine learning research. For example, NeurIPS began requiring ``impact statements'' in every submission to the conference in 2020; impact statements encourage authors to elucidate how their contribution might result in societal consequences, both positive and negative. Many of the publications we analyzed put forth some broad societal justification for the computer vision task they were contributing to, but did little to engage with the real world implications for that work. While several authors have put forth guidelines for writing impact statements (e.g., \cite{neuripsimpact}), we also encourage authors to incorporate their values and ethical considerations early on in the dataset construction process, before data collection, and into every step of the pipeline thereafter. Similarly, authors might outline refusals\,---\,data which they refused to collect, uses they refuse to condone, or opportunities for allowing data subjects or annotators to opt out\,---\,in their ethical statements \cite{garciaNoCriticalRefusal2020}.
\end{itemize}

\subsection{Model Work over Data Work}
\label{model-work}
Despite the fact that data is posited as crucial, many pieces of the data collection and curation process are missing from documentation, including, often, the data itself. Many datasets were not available at the time of writing, with URLs that go nowhere enshrined in archival papers. Reporting on the details of algorithms\,---\,or the model work\,---\,takes priority in the publications associated with these datasets. As discussed in Section \ref{data-as-essential}, the vast majority of these datasets do not get published unless they report some kind of algorithmic improvement to a machine learning task. Publications that report solely on datasets are typically not published. If they are published without a corresponding model or technical development, they are typically relegated to a non-archival technical report, rather than published in a top-tier venue. For this matter, reporting and evaluation of the model work is what is typically incentivized, rather than the careful, slow data work. Our findings here are in line with recent characterizations of machine learning dataset work as undervalued and deprioritized \cite{sambasivanEveryoneWantsModel2021, hutchinsonAccountabilityMachineLearning2020}.

Dataset documentation is itself undervalued in computer vision. The fact that the majority of papers in our corpus were focused on detailing the design of algorithms showcased that, while data is necessary to model development, it is auxiliary to the proposed model or method itself in the publishing process. Moreover, maintenance and stability of the data itself is devalued, with much of the data unavailable only a few years after being published. 

Even though most of the dataset creators report that they are interested in sharing the dataset as part of a larger research community, most are silent around the infrastructure necessary to maintain the upkeep of datasets. As noted above, while most of the datasets report an URL or have a find-able URL, many of the datasets do not have any kind of manner or mechanisms for data persistence. Practices such as providing each dataset with DOI, storing them at an institutional repository, or putting the dataset under version control are not followed or even mentioned within the majority of paper texts or on the websites of the datasets. As such, some datasets were no longer available, with sources in publications leading to expired links, making the data unavailable for use or scrutiny\,---\,even while it may be still in use by others who were lucky enough to download it when it was still available. The impermanency of datasets has several downstream effects, including increases in technical debt, lack of maintenance, and the inability to replicate and reproduce scientific results.

Information scholars note that data reuse itself is embedded in scientific practice and that "investments in data sharing… may have long-term consequences for the policies and practices of science" \cite{pasquettoReuseScientificData2017}. The amount of work that is put into data sharing has significant effects for the scientific work of communities of practice. Computer scientists have similarly written on the high cost of so-called "technical debt" in machine learning systems \cite{NIPS2015_86df7dcf}, including "data dependencies", which cost more than dependencies in machine learning code. Moreover, it's more appealing to create a new dataset than to maintain its upkeep, especially as datasets are created as the demands for new and more novel models emerge with increasing frequency. However, this leaves old, publicly available datasets in the lurch, to be deprecated through neglect and disuse.

The crisis of reproducibility in psychology and other empirical fields has become more pitched as large-scale empirical work has been adopted throughout computational sciences \cite{opensciencecollaborationEstimatingReproducibilityPsychological2015}. \citeauthor{stoddenEnablingVerificationComputational2018} found that for several issues of \textit{Computational Physics}, "Some artifacts [were] made available" in only 5.6\% of 306 cases. Moreover, "computer code, input and output data, with some reasonable level of documentation" was available for only 4\% of the cases, and that only 18\% of authors provided necessary artifacts for replication upon request \cite{stoddenEnablingVerificationComputational2018}. Similarly, \citeauthor{stoddenEmpiricalAnalysisJournal2018} found that, of 204 articles in the journal \textit{Science}, only 24 had the requisite information to obtain requisite code and data. After requesting code and data from authors, only 26\% of results in all articles were able to be replicated \cite{stoddenEmpiricalAnalysisJournal2018}. 

One of the datasets which meets the bar for doing the requisite data work -- in terms of data reuse, maintenance, and reproducibility -- is the CAFE dataset. CAFE was the only dataset hosted at an institutional repository (NYU Databrary) in our sample. In addition, being deposited in the Databrary also assigns a DOI to the dataset\footnote{http://doi.org/10.17910/B7301K}, which allows persistent access to the dataset.

\subsubsection{Recommendations for Incorporating the Value of Data Work}

We have several recommendations, in accordance with work on data curation and stewardship practices.

\begin{itemize}
    \item Assigning datasets a stable DOI and storage location would increase transparency, usability, and reproducibility. This may be in the form of a DOI, which will allow for a permanent reference to the dataset which is not bound to a particular URL, and will therefore be easier to find or link to from scientific papers and other persistent artifacts. It may also mean storing datasets in institutional repositories\,---\,such as Databrary, Harvard Dataverse, or Zenodo\,---\, such that data can be maintained at a persistent third-party location, rather than on a lab or personal website. Many of these services allow for version control of data, which is an added benefit, because as data changes and updates are made to datasets, new versions can be named and accessed, without erasing previous versions of the dataset. Scientific results can name a particular version of the data, rather than leaving would-be replicators to make assumptions about versioning.
    
    \item Create a data maintenance plan in order for data to remain relevant and useful. Dataset authors can maintain their data by ensuring it is still accessible. As \citeauthor{hutchinsonAccountabilityMachineLearning2020} state: "If you can't afford to maintain a dataset, and you also can't afford the risks of not maintaining it, then you can't afford to create it." Any updates should be clearly and transparently documented so that users understand how changes may impact their use. This would also allow researchers to understand that older versions of data may still be in use, and what those older versions look like. 
    
    \item Publishing rigorous and detailed dataset documentation alongside the dataset makes datasets more trustworthy, transparent, reproducible. Recognizing data work as a specialty area will increase incentives for documenting data more rigorously. Following \citeauthor{joLessonsArchivesStrategies2020}, dataset development should be understood as a speciality area of computer vision \cite{joLessonsArchivesStrategies2020}. It should be given intentional space in textbooks and curriculum, and publications focused exclusively on dataset developments should be recognized as meaningful contributions. We also suggest research tracks at top-tier computer science conferences that allow for data work to be incentivized as a publication in and of itself. On the level of machine learning more broadly, the inclusion of dataset documentation could be incentivized through journal policies and publication requirements. While publication policies alone are not sufficient to increase documentation sufficient for replication \cite{stoddenEmpiricalAnalysisJournal2018}, multiple frameworks exist for the reporting of data (e.g., \cite{gebruDatasheetsDatasets2020, hollandDatasetNutritionLabel2018, gil2016}). Our own open source codebook for this study provides a framework for dataset curators to build and document datasets.
    
\end{itemize}

\section{Conclusion}
Technical artifacts are imbued with politics. Previous scholars have examined the underlying politics and values of a variety of artifacts, many of which center data practices\,---\,from how values are shaped by the production of data \cite{vertesiValueDataConsidering2011} to how it then shapes our scientific practices more broadly \cite{bowker2005memory}. Researchers are increasingly interrogating the values of machine learning data, specifically, including how gender is represented in datasets for facial analysis tasks (e.g., \cite{scheuermanHowComputersSeeGender2019}), how racial bias in face data leads to biased outcomes (e.g., \cite{buolamwiniGenderShadesIntersectional2018}), and how images in object recognition datasets skew towards Western countries (e.g., \cite{shankarNoClassificationRepresentation2017a, Vries_2019_CVPR_Workshops}). We built on this work by broadly examining what the documentation and reporting practices of computer vision datasets say about the values of the discipline. We analyzed what aspects of dataset development authors gave space in the publications accompanying their datasets and how authors documented and described different components of the dataset. We also analyzed what went unsaid in these publications, in order to better understand what dataset developers valued. We found that, broadly, computer vision dataset practices value \emph{efficiency over care, universality over contextuality, impartiality over positionality,} and \emph{model work over data work}. For each of the silences we identified, we recommended potential steps dataset authors could take to attend to them. We hope that this move \,---\, acknowledging the values implicated in data creation and annotation, and taking steps to develop a careful, contextual, position-aware data practice \,---\, can lead to more replicable, accountable, and ethical research in computer vision, and machine learning more broadly. 



\begin{acks}
We'd like to thank Timnit Gebru, Ben Hutchinson, Vinodkumar Prabhakaran, and Madeleine Clare Elish for helping to craft elements of this project. We would also like to thank other members of the Genealogies of Data team -- Razvan Amironesei, Andrew Smart, and Hilary Nicole. A final thank you to Mally Dietrich and Katy Weathington for help with final copyedits.
\end{acks}

\bibliographystyle{ACM-Reference-Format}
\bibliography{main}

\received{January 2021}
\received[revised]{April 2021}
\received[accepted]{May 2021}

\end{document}